%% file: camera_ready_main.tex
\pdfoutput=1

\documentclass[11pt]{article}

\usepackage[final]{misc/acl}
\usepackage{listings}
\usepackage{times}
\usepackage{latexsym}
\usepackage{hyperref}

\usepackage[most]{tcolorbox}
\usepackage{tikz}
\usepackage{wrapfig}
\newtcolorbox{softbox}[1][]{
  enhanced,
  boxrule=0.2pt,
  colback=#1,
  arc=0.5mm,
  boxsep=2pt,
  left=5pt,
  right=5pt,
  before skip=6pt,
  after skip=6pt,
}
\newenvironment{dialoguebox}[1][]{
  \begin{figure*}[h!] 
  \centering
  \begin{tcolorbox}[
    enhanced,
    sharp corners=south,
    colback=gray!5!white,
    colframe=black!60,
    fonttitle=\bfseries,
    title=#1,
    boxrule=0.6pt,
    arc=4pt,
    left=4mm,
    right=4mm,
    top=2mm,
    bottom=2mm,
    width=\textwidth  
  ]
  \ttfamily
}{
  \end{tcolorbox}
  \end{figure*}
}

\newcommand{\teacher}{\textbf{Teacher:} }
\newcommand{\student}{\textbf{Student:} }

\usepackage[table]{xcolor}
\definecolor{Blue1}{RGB}{153, 204, 255}
\definecolor{Blue1}{rgb}{0.1216, 0.4667, 0.7059}
\definecolor{Pink1}{RGB}{219, 48, 122}
\definecolor{lred}{HTML}{ff9aa2}
\definecolor{lgreen}{HTML}{b5ead7}
\definecolor{dred}{HTML}{994f55}
\definecolor{dgreen}{HTML}{5c9c87}
\usepackage{colortbl,xcolor}

\usepackage[T1]{fontenc}
\usepackage{tikz}
\usepackage{subcaption}

\usepackage[utf8]{inputenc}

\usepackage{microtype}

\usepackage{inconsolata}
\usepackage{booktabs}
\usepackage{tabularx}

\usepackage{lscape}

\usepackage{xcolor}
\definecolor{GPTColor}{HTML}{E2F0CB}
\definecolor{LlamaColor}{HTML}{B5EAD7}
\definecolor{GeminiColor}{HTML}{C7CEEA}

\usepackage{rotating}
\usepackage{float}
\usepackage{times}
\usepackage{latexsym}
\usepackage{tablefootnote}
\usepackage{graphicx}
\usepackage{booktabs}
\usepackage[utf8]{inputenc}
\newcommand{\ignore}[1]{}
\newcommand{\squishlist}{
 \begin{list}{$\bullet$}
  { \setlength{\itemsep}{0pt}
     \setlength{\parsep}{2pt}
     \setlength{\topsep}{2pt}
     \setlength{\partopsep}{0pt}
     \setlength{\leftmargin}{1em}
     \setlength{\labelwidth}{1em}
     \setlength{\labelsep}{0.4em} } }

\newcommand{\squishend}{
  \end{list}  }

\usepackage[compact]{titlesec}
\usepackage{url}
\usepackage{cleveref}

\usepackage{tabularx}
\usepackage{geometry}
\usepackage{setspace}
\usepackage{colortbl}
\usepackage{xcolor}
\definecolor{darkgreen}{RGB}{0,100,0}
\usepackage{siunitx}  

\usepackage{array,multirow,multicol}
\usepackage{xspace}

\usepackage{enumitem}
\setlist{left=0mm,noitemsep}
\newcommand{\systemname}[0]{PATS\xspace}
\newcommand{\strategiser}[0]{Strategizer}
\newcommand{\responder}[0]{Responder}

\definecolor{studentcolor}{RGB}{0, 102, 204}

\title{PATS: Personality-Aware Teaching Strategies \\with Large Language Model Tutors}

\author{
  \textbf{Donya Rooein\textsuperscript{1}\thanks{Equal contribution}},
  \textbf{Sankalan Pal Chowdhury\textsuperscript{2}\footnotemark[1]},
  \textbf{Mariia Eremeeva\textsuperscript{2}}, \\
  \textbf{Yuan Qin\textsuperscript{3}},
  \textbf{Debora Nozza\textsuperscript{1}},
  \textbf{Mrinmaya Sachan\textsuperscript{2}},
  \textbf{Dirk Hovy\textsuperscript{1}}
\\
  \textsuperscript{1}Bocconi University \\
  \textsuperscript{2}ETH Zurich \\
  \textsuperscript{3}University of Zurich\\
 \small{
   \textbf{Correspondence:} 
   \href{mailto:donya.rooein@unibocconi.it}{donya.rooein@unibocconi.it}
 }\\
 \small{
 \includegraphics[height=1em]{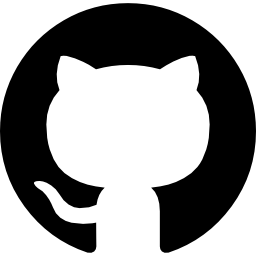} \href{https://github.com/donya-rooein/PATS}{github.com/donya-rooein/PATS}
 }
}

\begin{document}
\maketitle
\begin{abstract}
Recent advances in large language models (LLMs) demonstrate their potential as educational tutors. However, different tutoring strategies benefit different student personalities, and mismatches can be counterproductive to student outcomes. Despite this, current LLM tutoring systems do not take into account student personality traits.
To address this problem, we first construct a taxonomy that links pedagogical methods to personality profiles, based on pedagogical literature. We simulate student-teacher conversations and use our framework to let the LLM tutor adjust its strategy to the simulated student personality. 
We evaluate the scenario with human teachers and find that they consistently prefer our approach over two baselines. Our method also increases the use of less common, high-impact strategies such as role-playing, which human and LLM annotators prefer significantly. 
Our findings pave the way for developing more personalized and effective LLM use in educational applications.
\end{abstract}

\section{Introduction}
In most traditional classroom setups, a few students engage in interactive discussions, while many others stay silent. A common misconception is that quiet students are disengaged or not thinking critically, simply because they do not speak up. However, their silence may instead stem from differing personality types \cite{lawrence2015learning} and a mismatch between those and the teaching method. In fact, research in learning sciences indicates that students' preferred teaching strategies differ depending on their personality type \cite{murphy2017extraversion}, and that not all students benefit equally from all instructional approaches. For example, extroverts may prefer working with other students, while introverts may avoid active discussions and prefer to reflect before answering teachers’ questions.




\begin{figure}[h]
  \centering
  \includegraphics[width=\linewidth]{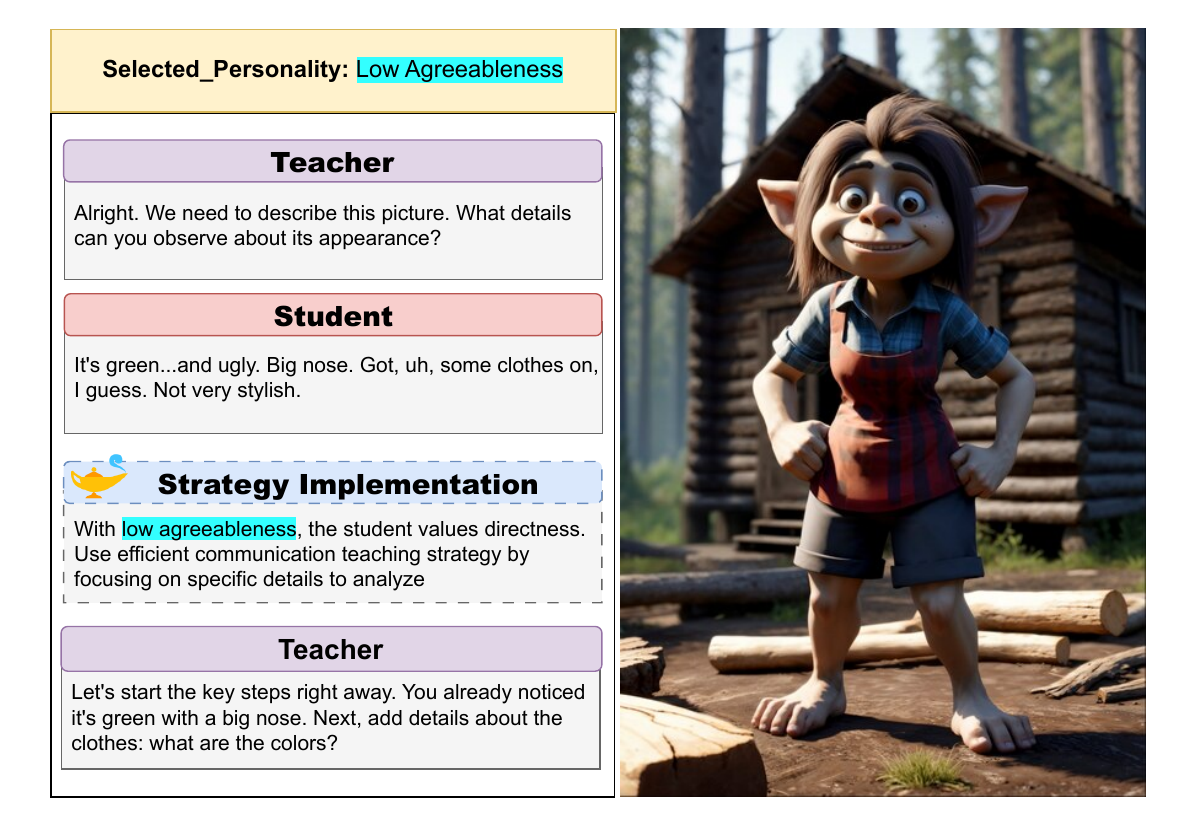}
  \caption{Teacher–student interaction in an image description task. The system identifies the student’s personality trait as \textsc{low agreeableness} and selects and implements a corresponding teaching strategy (e.g., efficient communication) to maintain the dialogue.}
  \label{fig:overview}
   \vspace{-0.9em}
\end{figure}


Large Language Models (LLMs) are majorly impacting personalized Intelligent Tutoring Systems (ITSs) \cite{jurenka2024towards}. LLM-based ITSs show potential in scaling up personalized tutoring, enhancing student learning outcomes, higher engagement, and offering personalized learning experiences \cite{liu2023novice,sonkar2023classdesignframeworkbuilding,Vanzo2024GPT4AAA,openai2025education, Scarlatos_2025}.

Successful personalized ITSs include MATHia\footnote{\url{https://www.carnegielearning.com/solutions/math/mathia/}}, AutoTutor\footnote{\url{https://autotutor.org/}}, and Squirrel AI\footnote{\url{https://squirrelai.com/}}, each employing distinct strategies for content personalization.
However, adaptive learning goes beyond mere content personalization \cite{rooein2022achat,graf2023exploring}; it also involves tailoring \textit{teaching strategies} to individual learners. For pedagogical ITSs to be effective, they must accommodate students with diverse personality traits \cite{zhang-etal-2025-simulating, robison2010developing}. \citet{liu-etal-2024-personality} propose a framework for constructing profiles of different student groups and demonstrate that LLMs are capable of personality-aware interactions in language learning contexts. However, their work does not explore whether, and how, LLMs can adapt their teaching strategies to align with specific personality needs, a topic that has been a longstanding focus in learning sciences research \cite{agler2020using, Conti2024TheAO, chu2025llm}. 

Those issues motivate our research question:
\vspace{-1.75mm}

\begin{tcolorbox}[colback=blue!0!white, colframe=blue!0!black, width=\columnwidth, boxrule=0.25mm, arc=0mm, auto outer arc, breakable]
\textbf{RQ}: Can \textit{LLM-based ITSs} personalize teaching strategies to better support learners with diverse personality profiles?
\end{tcolorbox}



\Cref{fig:overview} presents an overview of an LLM tutor that goes beyond simple question answering to incorporate more sophisticated teaching strategies. The system first detects the student’s personality traits from prior chat history and then applies an appropriate teaching strategy to generate the teacher’s responses. To ground this work, we use proprietary data from a real-world classroom study using LLM-based tutors to examine the diversity of students’ personality traits and how LLM tutors respond to their learning needs. 
However, LLM tutors primarily function as question-answering systems (see Section \ref{sec:school-example}), occasionally offering basic forms of scaffolding in their interactions with learners. They never tailor their strategy to student traits.

We create a taxonomy that maps teaching strategies to different personality traits, grounded in the learning sciences literature. Using this taxonomy, we propose a framework for incorporating personality-aware teaching strategies into LLM-based ITSs. To test the efficacy of our approach, we use LLM simulators. Compared to tests with human students, simulations are widely used and offer several advantages, enabling controlled experimentation with complex pedagogical setups, supporting a wider range of student-tutor interactions, providing a more time- and cost-effective approach, and posing fewer ethical challenges than tests involving sensitive data, such as personality tests \cite{markel2023gpteach,lu2024generative,zhang-etal-2025-simulating,martynova-etal-2025-llms}.

Our findings suggest that LLM-based tutors should be aware of students' personality traits and adjust their teaching strategies accordingly. Teachers and LLM annotators consistently preferred LLM-based teacher models that used explicit, personality-aligned teaching strategies to those that did not, especially in more engagement-sensitive dimensions such as motivation, empathy, and proactive support.

\paragraph{Contributions}
\begin{enumerate}
     \item We develop a taxonomy that links 17 teaching strategies to student personality traits.

    \item We use our taxonomy in simulations of two different educational tasks to generate personality-aware student-teacher conversations.
   
    \item We empirically evaluate the quality of simulated interactions with humans and LLMs and release the annotated dataset\footnote{This dataset is released under MIT license.}, along with a demo\footnote{The demo is available at \url{https://donya-rooein.github.io/PATS/demo.html}}, for future research.

\end{enumerate}

\section{Quantifying Student Personality Diversity in a Real Classroom}
\label{sec:school-example}
\citet{liu-etal-2024-personality} shows LLMs are positively perceived regarding personality simulation with the \textit{Big Five Theory} (\textbf{BFT}) \cite{McCrae1999five} for students. The BFT model with five traits\footnote{It consists of 44 items rated on a 1–5 scale.}, often called \textbf{OCEAN} includes: \textit{Openness} (the tendency to be curious, imaginative, and open to new experiences), \textit{Conscientiousness} (being organized, responsible, and goal-directed), \textit{Extraversion} (being outgoing, energetic, and sociable), \textit{Agreeableness} (being compassionate, cooperative, and trusting), and Neuroticism (the tendency to experience emotional instability, anxiety, and moodiness) \cite{John2008ParadigmST}. We aggregate personality traits based on low/high (i.e., the trait is high if scored $>3$ following the approaches in \citet{liu-etal-2024-personality}).
\citet{maurya-etal-2025-unifying} defines a taxonomy of essential pedagogical abilities for effective AI tutoring, but they mention only scaffolding in the context of teaching strategies. We analyze a real-life example of LLM integration in a classroom setting to highlight the diversity of student personalities and identify the limitations of LLMs as tutors in implementing teaching strategies.

\input{table/taxonomy}

We collaborated with the Minerva Research Association, an educational non-profit organization based in Vaud, Switzerland, to study the deployment of a GPT-based tutoring system within a language learning curriculum. In March 2025, seven language teachers from Instituto Pindemonte, a High School in Verona, Italy, were invited to integrate an LLM-based tutor into their classrooms, using the GPT-4o model from OpenAI \cite{openai2024gpt4technicalreport}. The LLM tutor used as a complementary support tool for homework, providing real-time feedback on grammar and argumentation in student essays.
During this deployment, a BFT survey was administered. We were provided with survey results and anonymized chat logs from 110 participating students (63 female, 45 male, 1 who preferred not to disclose, and 1 who identified as other), aged 16–21.\footnote{These chat logs are private; we were granted permission to publish only anonymized examples.} We identified the 10 most common personality profiles based on whether each trait scored above 3, and used these profiles in our analysis (see \Cref{appx:BFT-test}).
Despite the limited diversity in students' personalities, our qualitative analysis of 281 anonymized student–tutor chat logs revealed that the LLM tutor consistently relied on the same teaching strategy. It predominantly adopted a question–answering approach, regardless of the student’s personality, leading to relatively low student engagement (see the example in \Cref{appx:chat_example}).

Experiments with teachers and students in a real classroom provide the strongest validation. However, limited access to participants and time requirements prevented us from conducting a second round of human-subject experiments. Instead, we study the role of LLM simulators in modeling various teaching strategies. First, we need to design appropriate teaching strategies informed by the learning sciences. We then need to orchestrate the student and teacher models. The student model gets a personality-informed profile to produce responses consistent with its traits, while the teacher model adapts its instructional strategies accordingly.

\section{Taxonomy for Personality-Aligned Teaching Strategies}
\label{sec:taxonomy}
Learning science has shown that personality traits correlate with education-related factors such as learning approaches and learning outcomes \cite{jensen2015personality}. 
Following \citet{liu-etal-2024-personality} and \citet{agler2020using}, we use the BFT theoretical framework to model students' personality in tutoring conversations. 
Each trait in this personality framework is instantiated at two levels (low/high) and is associated with distinct behavioral attributes in learners.
We define a set of corresponding teaching strategies for each personality profile, grounded in research from the learning sciences (see Table \ref{tab:personality_strategies}). They are the basis for designing adaptive pedagogical responses in our LLM-powered ITSs.

By doing so, we also present the first comprehensive review of prior learning science literature that examines the relationship between student personality traits and effective teaching strategies \cite{BUSATO1998129,komarraju2005relationship,MASSA2006321,danesh2010relationship,harrington2010mbti,tlili2016role,agler2020using,zuo2024psychological}. We extract strategies that were originally designed for traditional classroom environments and included both individual and group-based interventions. Then, to ensure feasibility within LLM-powered ITSs, we filtered for the strategies that are implementable based on the following criteria: (a) they are suitable for one-on-one teacher-student interactions, (b) they can be implemented within educational tasks with LLMs, and (c) they are appropriate for the targeted educational level. Following these criteria, we excluded strategies that require the physical presence of other students, rely on group dynamics, or are heavily context-dependent (e.g., classroom management techniques or peer-mediated instruction in co-located settings). 
The results of our taxonomy are summarized in \Cref{tab:personality_strategies}, which maps teaching strategies to each learner's personality trait. 




In our taxonomy, students with \textit{High Openness} traits are inherently curious and creative. The most effective teaching approach for them is to provide real-world examples, following the Real-World Relevance strategy. Through this strategy, these students contextualize learning within meaningful scenarios. 
In addition, they benefit from the Task Variety (Variety) strategy because high-openness students are more likely to respond well to creative and varied learning methods \cite{komarraju2009role}.
In contrast, the literature suggests that students with \textit{Low Openness} prefer routine, structure, and familiarity, making Structured Direct Instruction (Instr.) and Step-by-Step Scaffolding (Scaff.) strategies more effective, which reduces discomfort and disengagement \cite{BUSATO1998129}. 
Learners with \textit{High Conscientiousness} are more likely to benefit from structured learning environments through Motivational (Motiv.) and progress-tracking (Progress) tasks that match their self-discipline and planning skills \citep{komarraju2009role}. For example, after completing a task, the tutor might ask questions such as \textit{``Can you explain how you decided on this answer?''}. 
Conversely, \citet{komarraju2005relationship} shows that \textit{Low Conscientiousness} learners often get distracted, requiring step-by-step supervision and breaking tasks into smaller subtasks. Additionally, incorporating short, low-stakes quizzes throughout the session as recommended by \citet{szpunar2013interpolated} can help reset attention and enhance cognitive engagement.

\citet{danesh2010relationship} characterizes \textit{High Extraversion} learners as individuals who have difficulty engaging with e-learning content independently, particularly in the absence of peers. Such learners are more responsive to a friendly tone than a mechanical one, and benefit from instructional strategies that incorporate role-play and empathetic prompts, such as \textit{``Imagine you’re helping a friend...''}. Those with \textit{Low Extraversion}, who may be reluctant, benefit from a calm, non-intrusive, Choice-Based Dialog (Choice) strategy, which reduces pressure while encouraging gradual verbal expression. \textit{High Agreeableness} often curious, open-minded, and engaged responds well to the Collaborative Peer Teaching strategy, which promotes responsibility \cite{agler2020using}. In contrast, \textit{Low Agreeableness} learners may exhibit disinterest or uncooperative behavior, benefiting more from structure and autonomy within clear boundaries \cite{komarraju2009role}.

Finally, learners with \textit{High Neuroticism}, who are often emotionally unstable or anxious, benefit from emotionally supportive learning environments that reduce stress. Teachers can support these students by normalizing mistakes (e.g., \textit{``Mistakes help us grow. Want to try again?''}), breaking tasks into small, manageable steps, avoiding strict time constraints, and offering flexible pacing with reassuring prompts \cite{zuo2024psychological}. Those learners with \textit{Low Neuroticism} may be more capable of handling independent learning without added emotional support. In such cases, it is beneficial for teachers to provide autonomy by allowing students to choose topics, task order, or challenge level \cite{Firat2022exploring,tlili2016role}. For example, prompts such as \textit{``Pick your next topic''} or \textit{``Explore this on your own first, I’ll step in if needed''} can foster independence. While our taxonomy summarizes a set of teaching strategies grounded in the literature (\Cref{tab:personality_strategies}), our framework is intentionally flexible: educators can adapt or extend these strategies to their unique classroom contexts and pedagogical goals.

\vspace{-1mm}
\section{A Framework for Personality-Aware Teaching Strategies (PATS)}
\label{sec:system}
\vspace{-1mm}
\subsection{Student Model}
The student model simulates a learner interacting with the tutor and exhibits behavior consistent with the personality traits defined by the BFT model. We extend student model prompts by \citet{liu-etal-2024-personality} to more effectively capture conversational dynamics at the start and end of interactions. We also make targeted modifications based on empirical observations, accounting for behavioral changes in state-of-the-art language models. While \citet{liu-etal-2024-personality} also considered students' cognitive levels, they did not release the corresponding prompts. Since cognitive level is not a central focus of our work, we omit this dimension from our student model. The refined prompts are in \Cref{appex:student_model}.

\subsection{Teacher Model}
\subsubsection{\systemname}
The process of interacting with a student, like any dialog, can be split into two components - a \textit{planning} component that decides medium to long-term strategy, and a \textit{responding} component that decides the immediate next response \cite{he2024planning}. This two-task approach has previously been used to enhance the pedagogical quality of tutoring \cite{macina2023mathdial, wang2023strategize}. In this work, we apply the same approach to choose strategies based on the student's profile. \systemname\ uses two components - the \strategiser\ and the \responder. The \strategiser\ decides which pedagogical strategy is best given the student's personality and conversation history. The decision is made in 3 steps:
\begin{enumerate}
    \item \textbf{Personality Selection:} First, the \strategiser\ is asked to select which trait (from the Big 5 Traits) of the student's personality it would like to act upon next, based on the task description and dialog context. This is important because personality-specific strategies in the literature act on individual personalities rather than on their combinations. Furthermore, the Trait Activation Theory \cite{tett2021trait} posits that different facets of an individual's personality activate at different times depending on the situation. 
    \item \textbf{Strategy Selection:} Once the personality trait has been selected, we ask the \strategiser\ to pick which strategy or strategies the model wants to implement. We provide the model with a list of all potential strategies applicable to the selected personality and their descriptions (as listed in Table \ref{tab:personality_strategies}) and ask it to select one or more of these.  We don't force it to pick a single strategy because some strategies can be complementary and give better results when used together.
    \item \textbf{Strategy Implementation:} Finally, we ask the model to explain how the chosen strategy or strategies would be implemented in the current scenario. Defining the strategy before implementing it is inspired by the Plan-and-Solve \citep{wang2023plan} or Plan-Then-Execute \citep{He_2025} framework, which have been found to be useful for prompting LLMs in prior work. The generated strategy is fed to the \responder\ as part of its prompt.
\end{enumerate}
The \responder\ handles the actual communication, implementing the strategies chosen by the \strategiser. It receives the task description, dialog context, student's personality, and strategy, including the implementation plan. The context comprises the last 4 utterances (2 each from the teacher and the student) and a summary of all previous turns (if there are more than 4 turns). This setup keeps the context length relatively consistent and brings it more in line with human working memory constraints \cite{van1983strategies}.  
Since a strategy can take more than one turn to be applied, the \strategiser\ checks whether the current strategy implementation is complete. If it determines that the strategy is complete, it can either end the conversation or select a new strategy. The end of the conversation consists of two final utterances, one asking the student to summarise what they learnt, followed by one bidding them farewell. 
\subsection{Baselines}
We compare \systemname\ with the following baselines:
\begin{enumerate}
    \item \textbf{simple prompt}: This is a minimal model similar to the teacher model in \citet{liu-etal-2024-personality}. It is given a task description, some pedagogical instructions to prevent revealing the answer to the student, and some setting constraints in its system prompt, and asked to carry out the entire conversation independently. Since it doesn't provide any personality information, it must either detect the student's personality based on their responses or disregard that aspect entirely. Despite the low amount of information, this baseline is important because it is how LLM tutors behave by default \citep{Vanzo2024GPT4AAA, alarbi2024making}.
    \item \textbf{personality prompt}: This model, in addition to all the information provided to \texttt{simple prompt}, gets a description of the student's personality identical to the one given to the student model. Since student personality is also an input to \systemname, this setup enables a fairer comparison. This baseline simulates an LLM tutor who knows the student's personality, but it isn't instructed with appropriate teaching strategies.
\end{enumerate}

\subsection{Tasks}
We set up two tasks to evaluate \systemname. In both tasks, a teacher interacts with a third-grade student, prompting them to discuss a subject. The goal of the interaction is to improve students’ language skills by engaging them in description, discussion, and interpretation. The subject depends on the task:
\begin{itemize}
    \item \textbf{Image Description:} This task involves describing cartoon-style images. We use AI-generated illustrations obtained from the Freepik platform\footnote{We collected images using the keywords ``Fantasy cartoon characters'' from \url{freepik.com}}.
    \item \textbf{Storytelling:} This task focuses on writing children’s stories with moral lessons\footnote{We collected stories from \href{https://ourlittlejoys.com/blog/moral-stories-for-kids/}{ourlittlejoys.com}}. Each story includes a title, the main text, and its moral.
\end{itemize}

We use 10 stories and 10 images as our subjects. For each subject, we generate a conversation for every possible personality with both our baselines and \systemname\ as the teacher model. Since there are $5$ personality traits, with $2$ possible values for each, we end up with $32$ personality profiles. Therefore, we have a total of $2\times32\times10=640$ dialogs for \systemname, and another $640$ for each of the baselines (total of $1920$ dialogs for each subject). The student model is held the same across all interactions (see \Cref{appex:student_model} for the student model). For evaluation, we run LLM assessments for all student profiles in \systemname. Then we perform fine-grained human evaluations on a subsample of dialogs.

\subsection{Evaluation Process}
\label{sec:evaluation}

\paragraph{LLMs Evaluation}

We run all comparisons as pairwise preferences, i.e, given a pair of conversations on the same subject with the same personality student, one carried out by \systemname, and the other by one of the baselines, which one is preferred. 
We test 3 state-of-the-art LLMs, namely GPT4o \cite{openai2024gpt4technicalreport}, Llama V3.3 70B \cite{dubey2024llama} and Gemini-2.0-Flash \cite{team2023gemini} evaluate our dialogs. We provide the LLMs with a description of the task and ask them to indicate their preference (see \Cref{sec:LLM-judge-prompt} for the full prompt). The complete dialog for the two systems is presented, labeling them as ``System 1'' and ``System 2''. The LLM first gives a chain-of-thought and then picks which system it thinks was better\footnote{It could also say that ``Both are Equal''}.

\paragraph{Human Evaluation}
We recruited expert teachers who are fluent in English and hold higher-education degrees. Each annotator was presented with a series of pairwise dialogue comparisons between \systemname and a baseline system. We subsampled $2$ baselines $\times$ $2$ tasks $\times$ top-$10$ personalities, resulting in $40$ dialogue pairs, which were then annotated by $4$ expert teachers (see \Cref{tab:teacher-demographics} for demographic details and \Cref{appx:study_experts} for instructions). Each teacher annotates $30$ dialogues, and each dialogue receives annotations from $3$ teachers. Annotators rank dialogs across \textit{five} dimensions: \textit{Motivation \& Support}, \textit{Suitability to Personality}, \textit{Proactivity}, \textit{Engagement}, and \textit{Empathy}. \Cref{tab:mapping-ques-stra} shows the dimensions we defined, based on their direct connection to teaching strategies.

\section{Results}
\label{sec:results}
In this section, we start by analysing teaching strategy distributions across \systemname dialogs. Then, we investigate the role of the proposed taxonomy in adapting the strategy to the personality trait selected by the teacher model. Finally, we report the results of both LLM and human evaluations of \systemname in comparison to baselines.

\begin{table}[t]
\centering
\small
\setlength{\tabcolsep}{6pt}
\renewcommand{\arraystretch}{1.1}
\begin{tabular}{lcc|lcc}
\hline
Strategy & Freq. & (\%) & Strategy & Freq. & (\%) \\
\hline
Motiv.     & 301 & \cellcolor{blue!48}38.4 & Relevance & 17 & \cellcolor{blue!12.2}2.2 \\
Scaff.     & 167 & \cellcolor{blue!31}21.3 & Instr.    & 13 & \cellcolor{blue!11.7}1.7 \\
Role-play  &  92 & \cellcolor{blue!22}11.7 & Choice    & 12 & \cellcolor{blue!11.6}1.6 \\
Friendly   &  84 & \cellcolor{blue!21}10.6 & Gamify    &  8 & \cellcolor{blue!11.1}1.1 \\
Comm.      &  54 & \cellcolor{blue!17}6.8   & Variety   &  3 & \cellcolor{blue!4}0.4 \\
Solo First &  29 & \cellcolor{blue!14}3.8   & Progress  &  3 & \cellcolor{blue!4}0.4 \\
\hline
\end{tabular}
\caption{Frequency and percentage of strategies in \systemname. Darker shading indicates higher prevalence.}
\label{tab:strategy-usage}
\end{table}

\begin{figure}[h]
\vspace{-2mm}
    \centering
    \includegraphics[width=1\linewidth]{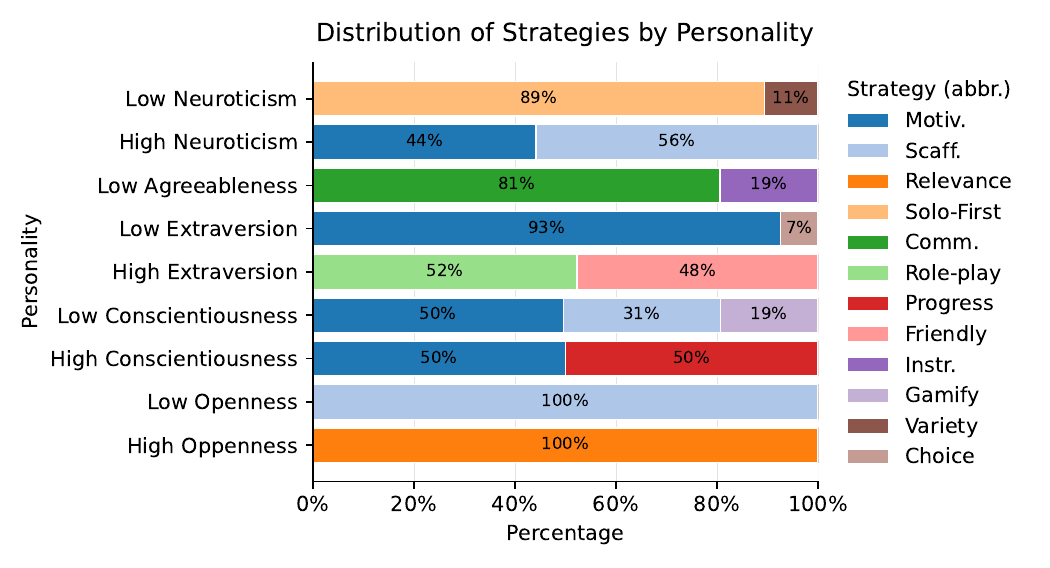}
    \vspace{-9mm}
    \caption{Distribution of teaching strategies used for each personality trait. During the analysis, \systemname\ never picked ``High Agreeableness''. }
    \label{fig:freq-st-trait}
    \vspace{-2mm}
\end{figure}

\begin{figure*}[th!]
        \centering
\includegraphics[width=0.9\linewidth]{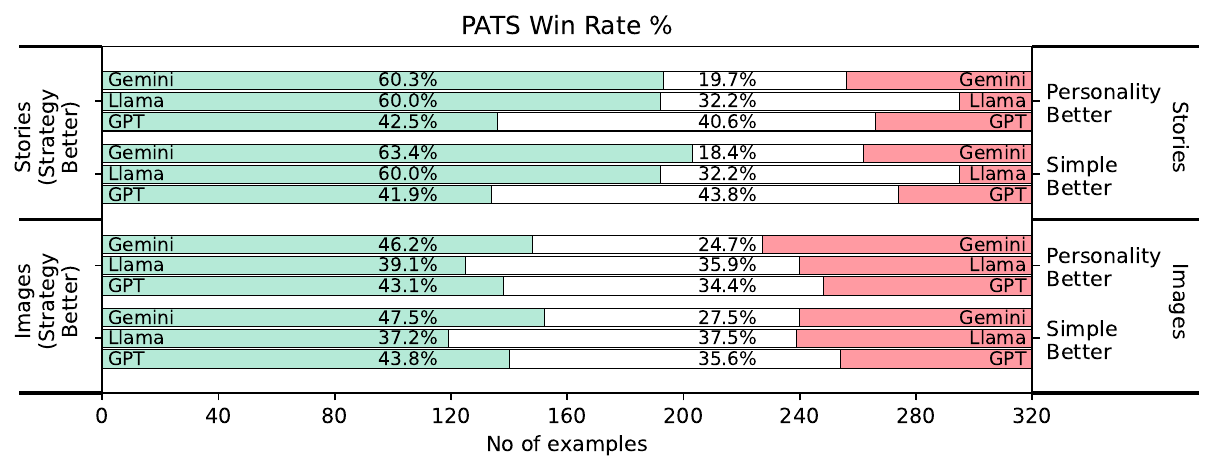}
\vspace{-3mm}
    \caption{Win rates for PATS over baselines on Image and Story tasks by LLM annotations (GPT-4o, Llama V3.3 70B, Gemini 2.0 Flash) across 10 runs and all 32 student profiles. Significance via one-sided t-tests shows that all 3 models find \systemname\ to be significantly better than both baselines. The white areas in the middle indicate ties.}
    \label{fig:llm_comp}
\end{figure*}
\begin{figure*}[t]
    \centering
    \begin{minipage}{0.48\linewidth}
        \centering
        \includegraphics[width=\linewidth, trim=2mm 25mm 5mm 5mm]{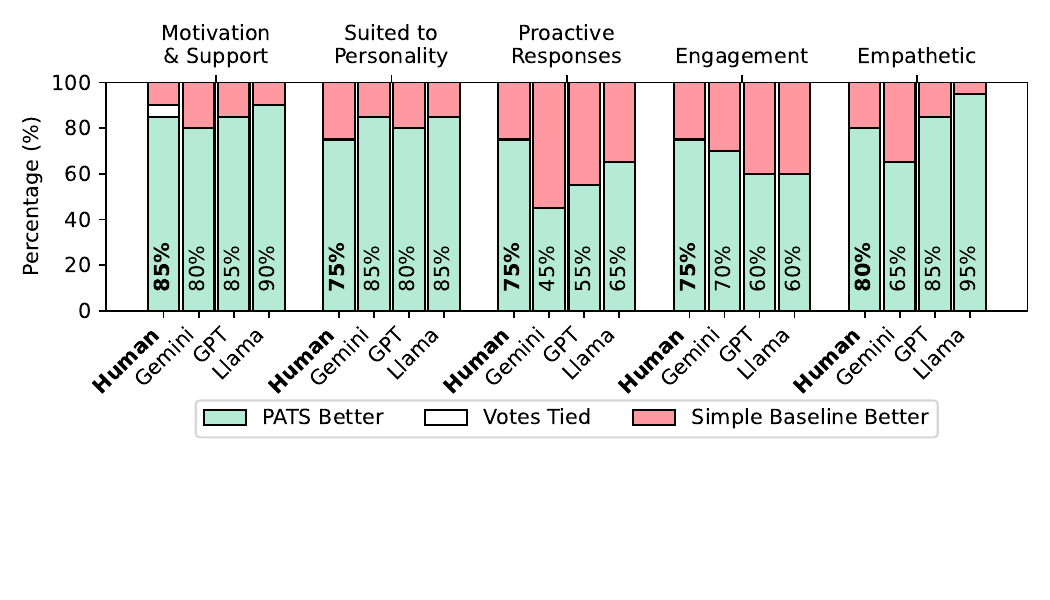}
        \vspace{-5mm}
        \subcaption{Win rates over \textbf{simple} baseline.}
        \label{fig:l1_expert}
    \end{minipage}\hfill
    \begin{minipage}{0.48\linewidth}
        \centering
        \includegraphics[width=\linewidth, trim=5mm 25mm 2mm 5mm]{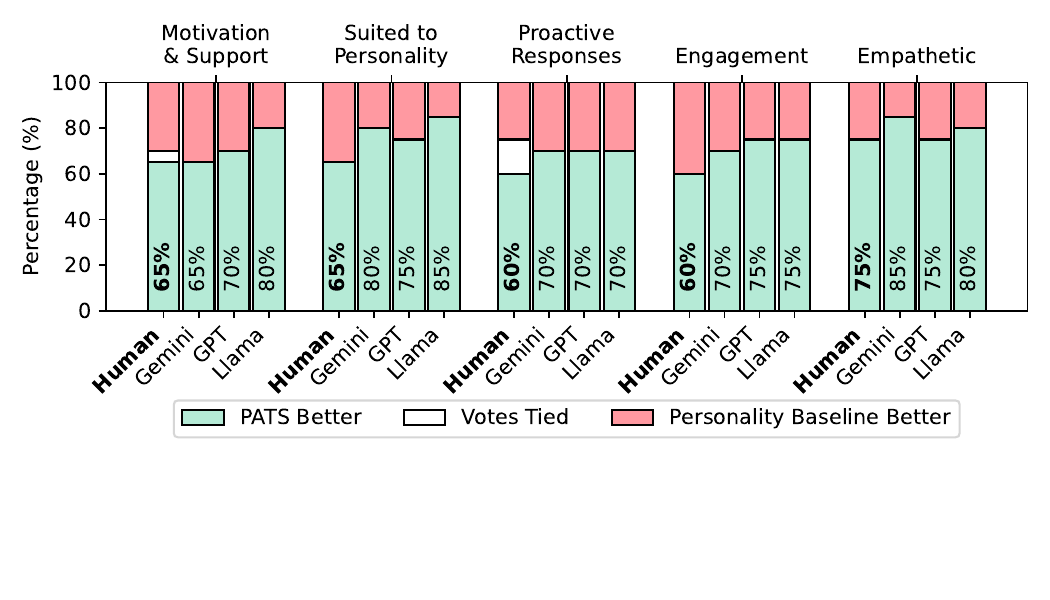}
        \vspace{-5mm}
        \subcaption{Win rates over \textbf{personality} baseline.}
        \label{fig:l2_expert}
    \end{minipage}
    \vspace{-2mm}
    \caption{Win rates across various dimensions for PATS over simple baseline (left) and personality baseline (right), ranked by LLMs (GPT-4o, Llama V3.3 70B, Gemini 2.0 Flash) and expert annotators. Each baseline has 20 instances, annotated three times using majority voting. Human preferences are highlighted in bold.}
    \vspace{-5mm}
    \label{fig:expert}
\end{figure*}
\subsection{Effectiveness of Teaching Strategy Implementation}
First, we evaluate whether the targeted teaching strategies are actually incorporated in \systemname generated dialogs. To this end, we randomly selected a sample of 100 dialogs and annotated them by three of the authors to determine the presence or absence of the specified strategies. The success rate is defined by whether the returned model's strategy implementation outputs mention and explain any selected strategy in relation to the student's personality trait. Across both tasks, three annotators achieved an overall average success rate of 94.6\% in implementing one of the intended teaching strategies. The average Cohen’s kappa scores for agreement between annotators across both tasks are 0.610.
This suggests that the model follows the intended instructional behavior in the vast majority of cases, demonstrating strong alignment with the specified strategies with respect to the selected personality traits.  \Cref{tab:strategy-usage} shows the frequency of strategy usage in \systemname dialogs. The results indicate that motivation and scaffolding strategies are most frequently employed by \systemname. This finding aligns with previous work by \citet{zhang-etal-2025-simulating}. In addition, \Cref{fig:freq-st-trait} illustrates the proportion of each strategy used across different personality traits. We observe that the teacher model selected the \textit{Relevance} strategy exclusively for learners with high openness, which aligns with the proposed taxonomy in \Cref{tab:personality_strategies}. Notably, the model never employed \textit{task variety} as a strategy. For low openness, although both \textit{Scaffolding} and \textit{Direct Instruction} were viable options, the model consistently chose only \textit{Scaffolding}, suggesting a selective preference that may reflect implicit biases of LLMs to perform scaffolding by default. Overall, the teacher model incorporates only 12 of the strategies defined in \Cref{tab:strategies_description}, which can be due to the limited number of student–teacher interactions collected in our dataset.

\subsection{Dialogue Level Comparison}
\Cref{{fig:llm_comp}} shows the preferences of three LLM annotations across all the simulated dialogs. To avoid order biases, we evaluate each LLM twice, flipping the order of the dialogs. When reporting LLM preferences, if flipping the order changes the preference, we report it as a tie. All LLMs significantly favor \systemname\ for both tasks and both baselines. In particular, for the \textit{Story} task, we reach $p<0.001$, and for the \textit{Image} task, it is slightly more contested, with $p=0.019$ for \textit{simple prompt} wrt Llama, and $p<0.01$ for everything else. The inter-model agreements (Fleiss $\kappa$), treating flipped evaluations as separate models, lie between $0.35$ and $0.41$, indicating fair to moderate agreement. LLMs almost never picked the ``Both are Equal'' option\footnote{Only Gemini picked this option in 4 cases of the \textit{image} task. We consider these 4 cases to be in favour of the preference of the flipped counterpart in \Cref{fig:llm_comp}.}.


To supplement the AI experiments, firstly, we asked educators hired from Prolific\footnote{\url{https://www.prolific.com/}} to compare $200$ dialogs. While dialogs from \systemname were preferred over both baselines ($p<0.02$ in both cases), the inter-annotator agreement was low ($\kappa<0.2$). Moreover, the proportion of total crowd workers passing the attention checks was 42\%, casting doubt on the reliability of these results\footnote{Details of this preliminary experiment are in \Cref{appx:study}}. We therefore conducted a smaller but more fine-grained study with four annotators.

We hired four expert teachers, carefully selected through professional educator channels. These teachers rate 40 pairwise comparison dialogs against each baseline.  \Cref{fig:expert} presents the results of these expert annotations alongside the annotations from three selected LLMs on the same questions. \Cref{fig:l1_expert} compares PATS against the simple baseline, and \Cref{fig:l2_expert}  compares PATS against the personality baseline, across five evaluation dimensions (Motivation \& Support, Suited to Personality, Proactive Responses, Engagement, and Empathetic). In all dimensions, humans and LLMs vote on PATS dialogs. Against the simple baseline, PATS achieved large gains in Motivation \& Support (85–90\%) and Suited to Personality (75–85\%), and showed consistent advantages in Engagement (60–75\%) and Empathy (65–90\%). The Proactivity of the teacher model shows more variance, with experts assigning PATS 75\% wins but LLMs ranging from 45–65\%. When compared to the personality baseline, PATS still outperformed across all dimensions. These results suggest that PATS not only surpasses a simple unconstrained tutor but also provides improvements over a personality-aware baseline by using teaching strategies.

\section{Related Work}
\paragraph{NLP in Education} Recent NLP advances have transformed ITSs from structured dialog systems \cite{GRAESSER199935, VanLehn2010,10.1145/3308560.3314191} into LLM-powered platforms that generate pedagogical content and enable free-form interactions \cite{schmucker2024rufflerileyinsightsdesigningevaluating, heyueya2024evaluatingoptimizingeducationalcontent, rooein-etal-2024-beyond, sonkar2023classdesignframeworkbuilding}, shifting the emphasis from static content to dynamic interaction \cite{Bernacki2021, li2024bringinggenerativeaiadaptive}.
However, it remains unexplored whether LLM tutors can adapt their pedagogical strategies given learners’ personality cues.

\paragraph{Personality-aware ITSs}
Personality modeling in educational systems typically relies on established psychological frameworks such as the BFT \cite{big5}, MBTI \cite{16pers}, and cognitive models like Kolb \cite{kolb}, and VARK \cite{vark}. 
State-of-the-art LLMs are capable of expressing controllable personas by generating trait-consistent responses \cite{jiang2024personallminvestigatingabilitylarge, pan2023llmspossesspersonalitymaking, zylowski-wolfel-2023-nlp, hu-collier-2024-quantifying, li-etal-2025-big5}. Methods such as prompt engineering \cite{jiang2024personallminvestigatingabilitylarge} or latent-space steering \cite{weng2024controllmcraftingdiversepersonalities} allow shifting the traits at inference time. 
Structured pipelines for generating and validating virtual students 
\cite{ma2024studentsexpertsnewai, huang2024orcaenhancingroleplayingabilities, yeo-etal-2025-pado} and dynamic-trait models that capture personality evolution \cite{pal2024discretepersonaspersonalitymodeling} are used to enhance personality simulations.
Personality simulations enable teachers to refine pedagogical strategies \cite{jin2025teachtunereviewingpedagogicalagents, Scarlatos_2025, puech-etal-2025-towards}, while for students, it boosts engagement and learning effectiveness \cite{ liu-etal-2024-personality, liu2025sizedoesntfitall}. 

\paragraph{Teaching strategies} 
The learning sciences describe teaching strategies such as direct instruction, worked examples, problem-based learning, and Socratic scaffolding \cite{rosenshine2012principles, Barrows, vandePol2010}.
Personality traits correlate with students' learning approaches and the benefits they derive from them \cite{BUSATO1998129, BIDJERANO200769}, also impacting the learning and interaction style preferences \cite{vark, felder}, as well as motivation, engagement, and achievement \cite{KOMARRAJU2011472, persinlear, agler2020using}. Consequently, a single pedagogical method rarely suits all learners.

\section{Conclusion}
LLMs are widely used for tutoring in education tasks, but do not adapt teaching strategies to students’ personality traits. Addressing this gap, we first map teaching strategies to personality traits, which we use to simulate student-teacher conversations by orchestrating multiple LLMs. We evaluate the simulated dialogs with humans and LLMs. \systemname uses a broader range of teaching strategies in our tutoring tasks, which is favored by human and LLM annotators. Our findings suggest that LLMs can and should be better aligned with pedagogical teaching strategies in tutoring tasks. This result underscores the importance of embedding pedagogical theory into LLM design for more effective and personalized educational support. 
However, we also find that many teaching strategies recommended in learning sciences are not easily implemented by prompt-only approaches, but require physical classroom presence. 

\section*{Acknowledgment}
We thank Alessandro Vanzo, Andrea Toffaletti and Lara Quarti from Minerva Research Association for giving us access to their ongoing research in LLM-based tutoring.

Donya Rooein, Debora Nozza, and Dirk Hovy are members of the MilaNLP group and the Data and Marketing Insights Unit of the Bocconi Institute for Data Science and Analysis. Donya Rooein and Dirk Hovy's research is supported through the European Research Council (ERC) under the European Union’s Horizon 2020 research and innovation program (No. 949944, INTEGRATOR). Debora Nozza's research is from the ERC under the European Union’s Horizon 2020 research and innovation program (grant agreement No. 101116095, PERSONAE). Sankalan Pal Chowdhury is partly funded by the ETH-EPFL Joint Doctoral Program for Learning Sciences.

\section*{Limitations}
This work proposes a taxonomy for implementing more complex teaching strategies tailored to student personality traits and evaluates an LLM-based framework for simulating student-teacher interactions. All data used in this work are intended solely for academic research purposes and should not be used outside of this context. Our student model is based on a single-personality framework, following prior work, and we do not incorporate other important student factors or needs. To rigorously assess the effectiveness of strategy-aligned dialogues, broader evaluation is needed, particularly through collaboration with educators and schools to examine long-term impacts on student learning outcomes. In addition, we acknowledge that conducting intervention studies in educational settings is a challenging and resource-intensive endeavor.

\section*{Ethical Considerations}
Our research with human annotators for both the prolific study and expert annotation received approval from the relevant Institutional Review Board (IRB). The personality tests conducted on students were part of an ongoing research project by a partner non-profit company, which has its own approval. They allowed us to use the anonymized data without publishing the full chat history. The details of our IRB approval and those of the partner are being withheld due to concerns about anonymity.
While the use of simulated student personalities offers a powerful tool for exploring adaptive pedagogical strategies, it is important to recognize the limitations of these models. Simulations, though valuable for research and development, may not fully capture the complexity and individuality of real learners. To ensure responsible application, developers and educators must remain mindful that LLM-generated feedback is not based on genuine understanding. Clear communication about the model’s capabilities and boundaries can help maintain appropriate levels of trust and ensure that LLMs are seen as support tools rather than infallible ones.
On a broader level, the integration of LLMs into education holds great promise for expanding access to personalized learning experiences, particularly in underserved areas. These systems can augment the efforts of human educators, offering scalable support where teacher resources are limited. However, thoughtful deployment is crucial to ensuring that LLMs complement, rather than replace, the human aspects of teaching, such as empathy and contextual judgment. By grounding implementation in ethical frameworks and educational best practices, we can harness the benefits of LLMs while preserving the irreplaceable role of teachers and promoting learner-centered outcomes.


\bibliography{bibliography.bib,anthology.bib}

\appendix

\section{Inference Details}
We use the following LLMs--- GPT4o (latest model from OpenAI API, queried April to May 2025), Llama V3.3 70B (hosted by \url{together.ai}), and Gemini-2.0-Flash (hosted by google ai studio). We collect all responses in English. We used default hyperparameters for all models. PATS dialogs generated by Gemini is licensed under the Creative Commons Attribution 4.0 License\footnote{\url{https://cloud.google.com/vertex-ai/generative-ai/docs/models/gemini/2-0-flash}}.

\paragraph{Computational Cost.} Despite being used for generating all dialogs, Gemini only cost us about \$12.00. GPT4o cost us about \$100.00, while Llama cost us about \$60.00. Generating a set of 32 dialogs takes about 20 minutes with either of the baseline models, and about 30 minutes with PATS. Note that this may vary considerably depending on server conditions. Evaluating dialog pairs takes about 15 minutes per 32 dialogs with Gemini, and 30 minutes per 32 dialogs with GPT4o and Llama. For all cases, the limiting factor is latency, not rate limit.

\paragraph{Packages for Evaluation.} Scipy (1.15.3) is used to calculate Pearson's Correlations and statistically significant Tests.

\section{Personality Test}
\begin{figure}
    \centering
    \includegraphics[width=1\linewidth]{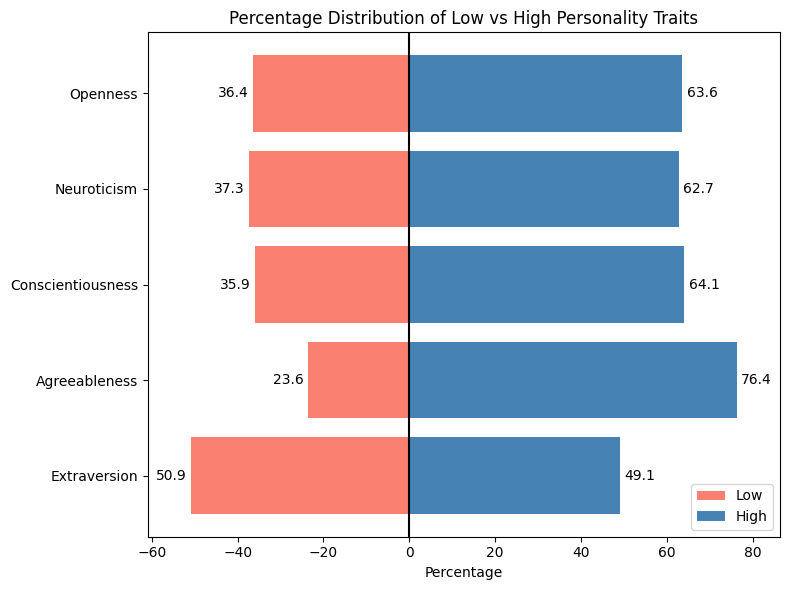}
    \caption{Percentage distribution of ``Low'' vs ``High'' personality traits among 110 students. We consider scores of $\ge3$ as \textit{High}, and $\le3$ as \textit{Low} in traits. }

    \label{fig:personality_test}
\end{figure}

Here we report the results of the personality traits in \Cref{tab:ocean_personality_profiles}. In addition, we report the percentage of low \ high for each trait in \Cref{fig:personality_test}.

\begin{table*}[]
\centering
\begin{tabular}{|c|c|cc|cc|}
\hline
 &  & \multicolumn{2}{c|}{\textbf{Image}} & \multicolumn{2}{c|}{\textbf{Story}} \\ \cline{3-6} 
\multirow{-2}{*}{\textbf{Model}} & \multirow{-2}{*}{\textbf{Personality}} & \multicolumn{1}{c|}{\textbf{F1 Score}} & \textbf{Ave. F1} & \multicolumn{1}{c|}{\textbf{F1 Score}} & \textbf{Ave. F1} \\ \hline
 & Openness          & \multicolumn{1}{c|}{\cellcolor{Blue1!33}\textbf{79.29\%}} &  & \multicolumn{1}{c|}{\cellcolor{Pink1!40}\textbf{83.15\%}} &  \\ \cline{2-3} \cline{5-5}
 & Conscientiousness & \multicolumn{1}{c|}{\cellcolor{Blue1!20}\textbf{68.25\%}} &  & \multicolumn{1}{c|}{\cellcolor{Pink1!27}\textbf{73.02\%}} & \\ \cline{2-3} \cline{5-5}
 & Extraversion      & \multicolumn{1}{c|}{\cellcolor{Blue1!44}\textbf{85.03\%}} & 71.18\% & \multicolumn{1}{c|}{\cellcolor{Pink1!40}83.03\%}  & 74.90\%  \\ \cline{2-3} \cline{5-5}
 & Agreeableness     & \multicolumn{1}{c|}{\cellcolor{Blue1!22}70.43\%}         &  & \multicolumn{1}{c|}{\cellcolor{Pink1!25}71.33\%}         &  \\ \cline{2-3} \cline{5-5}
\multirow{-5}{*}{GPT} & Neuroticism      & \multicolumn{1}{c|}{\cellcolor{Blue1!7}52.91\%}          &  & \multicolumn{1}{c|}{\cellcolor{Pink1!16}63.95\%}         &  \\ \hline
 & Openness          & \multicolumn{1}{c|}{\cellcolor{Blue1!31}77.03\%}         &  & \multicolumn{1}{c|}{\cellcolor{Pink1!22}69.44\%}         &  \\ \cline{2-3} \cline{5-5}
 & Conscientiousness & \multicolumn{1}{c|}{\cellcolor{Blue1!6}51.36\%}          &  & \multicolumn{1}{c|}{\cellcolor{Pink1!12}57.99\%}         &  \\ \cline{2-3} \cline{5-5}
 & Extraversion      & \multicolumn{1}{c|}{\cellcolor{Blue1!29}75.00\%}         & \textbf{74.86\%}  & \multicolumn{1}{c|}{\cellcolor{Pink1!60}\textbf{93.02\%}} & \textbf{78.26\%} \\ \cline{2-3} \cline{5-5}
 & Agreeableness     & \multicolumn{1}{c|}{\cellcolor{Blue1!60}\textbf{93.02\%}} &  & \multicolumn{1}{c|}{\cellcolor{Pink1!56}\textbf{90.06\%}} &  \\ \cline{2-3} \cline{5-5}
\multirow{-5}{*}{Gemini} & Neuroticism   & \multicolumn{1}{c|}{\cellcolor{Blue1!32}\textbf{77.89\%}} &  & \multicolumn{1}{c|}{\cellcolor{Pink1!38}\textbf{80.76\%}} & \\ \hline
\end{tabular}
\caption{Performance of GPT and Gemini for Student Model backend. Numbers in bold indicate the best performance for each personality trait across the models. Overall, the Gemini model has a better average F1 score across all personality traits.}
\label{tab:student}
\end{table*}

\label{appx:BFT-test}

\begin{table}[t]
\centering
\begin{tabular}{ccccc r}
\toprule
\textbf{O} & \textbf{C} & \textbf{E} & \textbf{A} & \textbf{N} & \textbf{Popularity} \\
\midrule
\rowcolor[HTML]{D8FBD9} H & H & L & H & H & 12.6\% \\
\rowcolor[HTML]{D8FBD9} H & H & H & H & H & 10.5\% \\
\rowcolor[HTML]{D8FBD9} H & H & H & H & L & 8.2\% \\
\rowcolor[HTML]{D8FBD9} L & H & L & H & H & 5.5\% \\
\rowcolor[HTML]{D8FBD9} H & L & H & H & H & 5.1\% \\
\rowcolor[HTML]{D8FBD9} L & H & H & H & H & 4.4\% \\
\rowcolor[HTML]{D8FBD9} H & L & L & H & H & 4.4\% \\
\rowcolor[HTML]{D8FBD9} L & L & L & H & H & 4.1\% \\
\rowcolor[HTML]{D8FBD9} H & L & L & H & L & 4.1\% \\
\rowcolor[HTML]{D8FBD9} H & H & L & H & L & 3.9\% \\
L & L & H & H & L & 3.6\% \\
L & H & L & H & L & 3.6\% \\
H & L & H & H & L & 3.2\% \\
H & H & H & L & L & 3.0\% \\
H & H & L & H & L & 3.0\% \\
L & L & L & L & H & 2.3\% \\
H & H & L & L & H & 2.3\% \\
L & L & H & H & L & 2.0\% \\
L & L & H & H & H & 1.9\% \\
L & H & H & L & H & 1.9\% \\
L & H & H & L & L & 1.6\% \\
H & H & H & L & H & 1.4\% \\
L & L & H & L & L & 1.4\% \\
H & H & L & L & H & 1.2\% \\
L & H & L & L & H & 1.2\% \\
L & L & L & L & L & 0.9\% \\
L & H & H & L & L & 0.7\% \\
L & L & H & L & H & 0.7\% \\
H & L & L & L & H & 0.5\% \\
H & L & L & L & L & 0.5\% \\
L & L & L & H & L & 0.5\% \\
H & L & H & L & L & 0.0\% \\
\bottomrule
\end{tabular}
\caption{Prevalence of student personality profiles based on the OCEAN model. Traits are coded as High (H) or Low (L). Students judged as borderline on a trait were split 50-50 into high (H) and low (L) groups. The top 10 personalities we used for our Human Experiments are highlighted in green.}
\label{tab:ocean_personality_profiles}
\end{table}

\section{Downside of No personalisation, an example}
\label{appx:chat_example}
We obtained access to this anonymized dataset through a collaborator in our research group, with the stipulation that it be used exclusively for research purposes. Here is an example of anonymized dialogs: Dialogs 1 and 2 show an example of how using the same style for students with different personalities can lead to issues. Both dialogs are about the same topic and question, but the students have different personalities. The student in Dialog 1 has High Conscientiousness, so they try to work with the tutor to get an answer. They do stop responding at one point (as can be seen by the consecutive teacher utterances), possibly because their openness is low and neuroticism is high. However, in Dialog 2, it is pretty clear that the student simply starts using ChatGPT (as seen by the following language, use of `•' for bullets, which is not present on a standard keyboard, and also verified by AI detection tools) and would effectively learn nothing. This is most likely a result of Low Conscientiousness. More Motivation and some Gamification could probably help here.
\input{generated/dialog1}
\input{generated/dialog2}

\begin{table}[ht]
\centering\small
\begin{tabular}{lp{2.5cm}}
\toprule
\textbf{Strategy} & \textbf{Abbreviation} \\
\midrule
Motivational Encouragement & Motiv. \\
Step-by-step Scaffolding & Scaff. \\
Friendly Tone & Friendly \\
Playful Role-play & Role-play \\
Real-world Relevance & Relevance \\
Efficient Communication & Comm. \\
Choice-based Dialog & Choice \\
Structured Direct Instruction & Instr. \\
Gamified Learning Items  & Gamify\\
Task Variety & Variety \\
Solo First Then Support & Solo-First\\
 Progress Tracking & Progress\\
\bottomrule
\end{tabular}
\caption{Abbreviations used for teaching strategies in the paper.}
\label{tab:strategy_abbreviations}
\end{table}

\section{Teaching Strategies}
\label{appex:teaching_strategies}
In \Cref{tab:strategies_description}, we describe our teaching strategies in the taxonomy together with references. The list of abbreviated strategies is available in \Cref{tab:strategy_abbreviations}.
\input{generated/taxonomy_table}

\section{Prompts and Design}
All prompts shown here are for the image task. For the story task. The image is passed in every user message unless otherwise stated. For the story task, the prompts are changed to reflect the change in task, and the story is passed in every system prompt.

\subsection{Student Model}
\label{appex:student_model}
We based our student model on \cite{liu-etal-2024-personality}. However, to account for the fact that the SOTA LLMs have changed since their work, we compared two SOTA LLMs at the time or our experiments, namely GPT4o and Gemini-2.0-Flash. \Cref{tab:student} shows the results. While there are other well-known LLMs, most of them do not support image inputs, making it impossible to carry out the image task with them. Similar to \cite{liu-etal-2024-personality}, we use GPT4o to evaluate personalities from the conversations. The scores obtained are shown in \Cref{tab:student}. As can be seen, Gemini has the higher scores on average, so we pick it as our student model backend.

Each student utterance is generated in two stages. First, we use the chat API of gemini with the following system prompt to generate 5 potential utterances:
\begin{quote}
[Role \& Task Definition] You are a primary school student. You are taking a language learning class, and describing the given picture.

[Personality Description]

\{personality\}

[Behavior Constraint] Always wait for the teacher's input before proceeding to the next step, unless your personality dictates otherwise. You may indicate body language cues like (looks away), (thinking), (stuttering), (confidently), etc., in brackets to show your emotions and feelings. YOU MAY NOT LEAVE THE CLASS
\end{quote}

Thereafter, a second prompt is used to select the best response. The image/story is not passed as we want to focus on the personality here. In this call, we use the completion API. The system prompt for this stage is as follows.

\begin{quote}
A student primary school student is interacting with a teacher, trying to describe a picture. The conversation so far has gone as follows:

\{history\}

We want to pick the next utterance that is most consistent with the student's personality. The student's personality, along with the definition of each personality trait is as follows:
\{personality\}

Given below are 5 potential next utterances for the student:

\{utterances\}

Pick the utterance that is most consistent with the student's personality. Explain your decision.
\end{quote}
This system prompt is used with two successive user prompts. The first prompt generates a chain of thought:

\begin{quote}
    Explain your reasoning for the choice you made.
\end{quote}
The second user prompt generates gets the model choice based on the chain of thought
\begin{quote}
    You have previously given the following answer and explanation:
    \{chain-of-thought\}
    
    Based on your explanation, now pick the best response. Only give a number between 1 and 5, no other explanation needed.
\end{quote}

The selected option is added to the history of the chat model and the cycle continues. This second stage was added to ensure that the model stays true to the student personality even as the context grows longer.
\subsection{Teacher Model Prompts}
\subsubsection{Baselines}
Both baselines use a single system prompt similar to the student. The remaining conversation is handled by the chat API of Gemini.
The simple prompt baseline uses the following prompt:

\begin{quote}
[Role \& Task Definition] You are a primary school language teacher. You teach the student to describe the picture. Try to get the student to come up with the description by gently nudging them, and avoid describing things yourself

[Pedagogical Instruction] You are using the knowledge construction approach to help me describe the picture. This involves any one of the following: building on prior knowledge, selecting information, integrating ideas, and making inferences.

[Behavior Constraint] Ask me only one question at a time. Always wait for my input before proceeding to the next step. Correct my answers if they are inaccurate.

[Setting Constraint] We are simulating verbal communication so keep utterances reasonably short, and without newlines. Once the classroom has been sufficiently described, for a 3rd-grade student, have the student summarise and end the conversation by <finish>.
\end{quote}
The personality prompt baseline uses a slightly modified version of the above prompt to add in the student personalities.
\begin{quote}
[Role \& Task Definition] You are a primary school language teacher. You teach the student to describe the picture. Try to get the student to come up with the description by gently nudging them, avoid describing things yourself

[Pedagogical Instruction] You are using the knowledge construction approach to help me describe the picture. This involves any one of the following: building on prior knowledge, selecting information, integrating ideas, and making inferences.

[Student Personality] From your previous interactions with the student, you know that the students personality, given by big 5 traits is as follows:

\{personality\}

Interact with the student in a way that is best for their personality.

[Behavior Constraint] Ask me only one question at a time. Always wait for my input before proceeding to the next step. Correct my answers if they are inaccurate.

[Setting Constraint] We are simulating verbal communication so keep utterances reasonably short, and without newlines. Once the classroom has been sufficiently described, for a 3rd-grade student, have the student summarise and end the conversation by <finish>.
\end{quote}
\subsection{\systemname}
\systemname\ makes use of only the completion api from Gemini. The system prompt for the \responder\ introduces the task and student personality. In the case of the story task, the system prompt includes the story. In case of the image task (for which we are giving the prompts) the image is passed with the user prompt as the system prompt does not accept images.
\begin{quote}
You are a primary school language teacher. You teach the student to describe the picture. Try to get the student to come up with the description by gently nudging them, avoid describing things yourself. From your previous interactions with the student, you know that the students personality, given by big 5 traits is as follows:

\{personality\}

Interact with the student in a way that is best for their personality.
\end{quote}

The user prompt for the \responder\ changes based on where we are in the conversation. For the first utterance there is no context, strategy or summary. The prompt is simply:
\begin{quote}
    Give the first utterance to start the conversation. Give only the first teacher utterance, do not prepend the phrase "Teacher:" to it
\end{quote}

For the next two teacher utterances, we include the entire conversation so far as context. At this point it also starts receiving a strategy

\begin{quote}
Because the student has \{persona\}, you have decided to pursue a strategy of \{strategy\}. \{turn\_count\} The conversation so far has proceeded as follows:

\{context\}

Give the next utterance to continue the conversation. Stick to the current strategy which is:

\{strat\_desc\}. 

Do not end the conversation but make sure to take into account the body language of the student. Give only the teacher utterance, do not prepend the phrase "Teacher:" to it.
\end{quote}

From the fourth utterance onwards, only the last $4$ utterances ($2$ from the tutor and $2$ from the student) are passed as context. The model also receives a summary of what the student has achieved in previous turns. The prompt for this stage is as follows:

\begin{quote}
Because the student has \{persona\}, you have decided to pursue a strategy of \{strategy\}. \{turn\_count\} 
The early part of the conversation can be summarised as follows:\{summary\}

The 4 most recent utterances are as follows:

\{context\}

Give the next utterance to continue the conversation. Stick to the current strategy which is:

\{strat\_desc\}. 

Do not end the conversation but make sure to take into account the body language of the student. Give only the teacher utterance, do not prepend the phrase "Teacher:" to it.
\end{quote}

The second-to-last utterance is supposed to ask the student to summarize the conversation as their last utterance. There is no other strategy here. That is generated using the following prompt:

\begin{quote}
    The early part of the conversation can be summarised as follows:\{summary\}
    
    The 4 most recent utterances are as follows:
    
    \{conversation\}
    
    The conversation has to be finished now. Give your penultimate utterance, asking the student to summarise all they have said about the picture so far.
    
    Give only the teacher utterance, do not prepend the phrase "Teacher:" to it.
\end{quote}

The last utterance simply bids farewell to the student and uses the following prompt:

\begin{quote}
    The early part of the conversation can be summarised as follows:{summary}
    
    The 4 most recent utterances are as follows:
    
    \{conversation\}
    
    the conversation has to be finished now. Give your final utterance bidding farewell to the student.
    
    Give only the teacher utterance, do not prepend the phrase "Teacher:" to it.
\end{quote}

The \textit{turn\_count} previously used in the user prompts is a single sentence telling how long the current strategy has been going on for. It either states ``This is the first turn of the current strategy.'' if this is a new strategy or says ``The current strategy is being worked on since last $x$ teacher turns'' where $x$ is the number of teacher turns since the start of the current strategy.

The summary is generated by its own pair of prompts. The user prompt contains the entire dialog so far. The system prompt is as follows:

\begin{quote}
    You are given a conversation between a teacher and a student. The student is trying to describe the picture. Please summarise the conversation in at most 50 words, focusing on what the student has already covered and the body language of the student
\end{quote}

The \strategiser\ is called first time before the second teacher utterance. Thereafter, it is called whenever the current strategy is deemed to have been completed. The system prompt for taking this decision is as follows:

\begin{quote}
A teacher is having a conversation with a student, helping them to describe a picture. The teacher is attempting to follow a specific strategy to help the student describe a picture. this strategy was chosen based on a personality trait of the student. Given the personality trait, the current strategy, all the utterances since the current strategy begun and a summary of the conversation so far, please decide if the strategy needs to continue, or if the teacher should move to a new strategy.
\end{quote}

\noindent which is paired with the following user prompt:

\begin{quote}
Summary of full conversation:

\{summary\}

Chosen Personality Trait:\{persona\}

Current Strategy:

\{strategy\}:\{strat\_desc\}

Conversation since start of strategy:

\{context\}

Please choose if we should move on to a new strategy, or continue the current one. Only say 'continue' or 'new'. No other text is required.
\end{quote}
The summary here is same as the summary for the \responder. For the first call, where there is no summary, we simply pass ``There is no summary, this given conversation is all of it''. The context length is variable depending on how long the current strategy has been going.

Every time the current strategy is deemed to have been completed, we also check if the overall conversation is completed. Here, the user prompt is once again just the entire conversation, while the system prompt is as follows:

\begin{quote}
You are a primary school language teacher. You teach the student to describe the picture. The conversation so far is given below. The conversation is considered finished if the student has described the picture and analysed it well enough for a 3rd grader. Is the conversation finished? If unsure, err towards saying "yes". Only say "yes" or "no". No other text is required.
\end{quote}

Finally, we come to the \strategiser, which uses three successive user prompts with the same system prompt. The system prompt is as follows:
 \begin{quote}
 A teacher is having a conversation with a student, helping them to describe the picture. The teacher wants to pick a strategy that works best for the students personality. To achieve this, we first pick one of the five personality traits of the student. then based on the personality trait, we pick a type of stratedy. Then we expand on how to follow the strategy.
 \end{quote}

 The first user prompt decides which personality to focus on based on the conversation so far:

 \begin{quote}
We start of by picking which personality trait of the student we want to focus on. The five personality traits of the student, with their identifying features are as follows:

\{persona\_choices\}.

The last few turns of the conversation have proceeded as follows:

\{context\}

Pick one of the five personalities to focus on.

Only say one letter, 'A' for Agreeableness, 'C' for Conscientiousness, 'E' for Extraversion, 'O' for Openness and 'N' for Neuroticism. No other text is required.
 \end{quote}

 The `persona\_choices' used here is different from the personalities used before. It just lists $5$ the five personality traits of the student one in each line, for example
 \begin{quote}
     High Agreeableness
     
     Low Openness
     
     High Conscientiousness
     
     High Extraversion
     
     Low Neuroticism
 \end{quote}
\noindent The selected personality is left in the same format to be used in other prompts.

Next we proceed to selecting strategy, which uses the following prompt:
\begin{quote}
    
Next, we pick the strategy. Once again, the last few turns of the conversation look as follows:

\{context\}

Your previously chosen personality trait is \{persona\}. For this personality, the following strategies are available 

\{strategies\}

Please pick on or more of the strategies. Only list the selected strategy names, separated by commas. No other text is required.
\end{quote}
\noindent Available strategies are selected based on persona in accordance with \Cref{tab:personality_strategies}. We also provide strategy descriptions alongside the choices based on \Cref{tab:strategies_description}

The final step is to decide how the chosen strategy(ies) would be implemented, which uses the following user prompt:
\begin{quote}
The next step is to decide how you want to implement the strategy. The last few turns of the conversation look as follows:

\{context\}

Your previously chosen personality trait is \{persona\}. Your previously chosen strategy(ies) is(are) \{strategy\}. Explain how this strategy can be used to help the student describe the picture in the next few turns. Keep your explanation short, no more than 50 words. Also, mention how many utterances you think this would take. This should be between 1 and 3
\end{quote}

\section{The dialog comparison study}

\subsection{Study 1: Prolific Teachers}
\label{appx:study}
\textbf{Annotator details}. We hired 120 annotators through Prolific, filtered for teachers who are fluent in English and hold a higher education degree. We paid £9.00 hourly. Annotators are instructed to read both conversations and answer the question: ``Which dialog provides a better learning experience overall?'' An option to select ``No Preference'' is also available, but discouraged unless the worker genuinely cannot decide.

To ensure reliability:
\begin{itemize}
\item Each pair is evaluated by at least 3 different annotators.
\item Dialog order is randomized to control for presentation bias.
\item We include a few control pairs with clearly better or worse responses to filter inattentive workers.
\end{itemize}

\subsubsection{Interface}
\label{interface}
\begin{figure*}[h!]
    \centering
    \includegraphics[width=\textwidth]{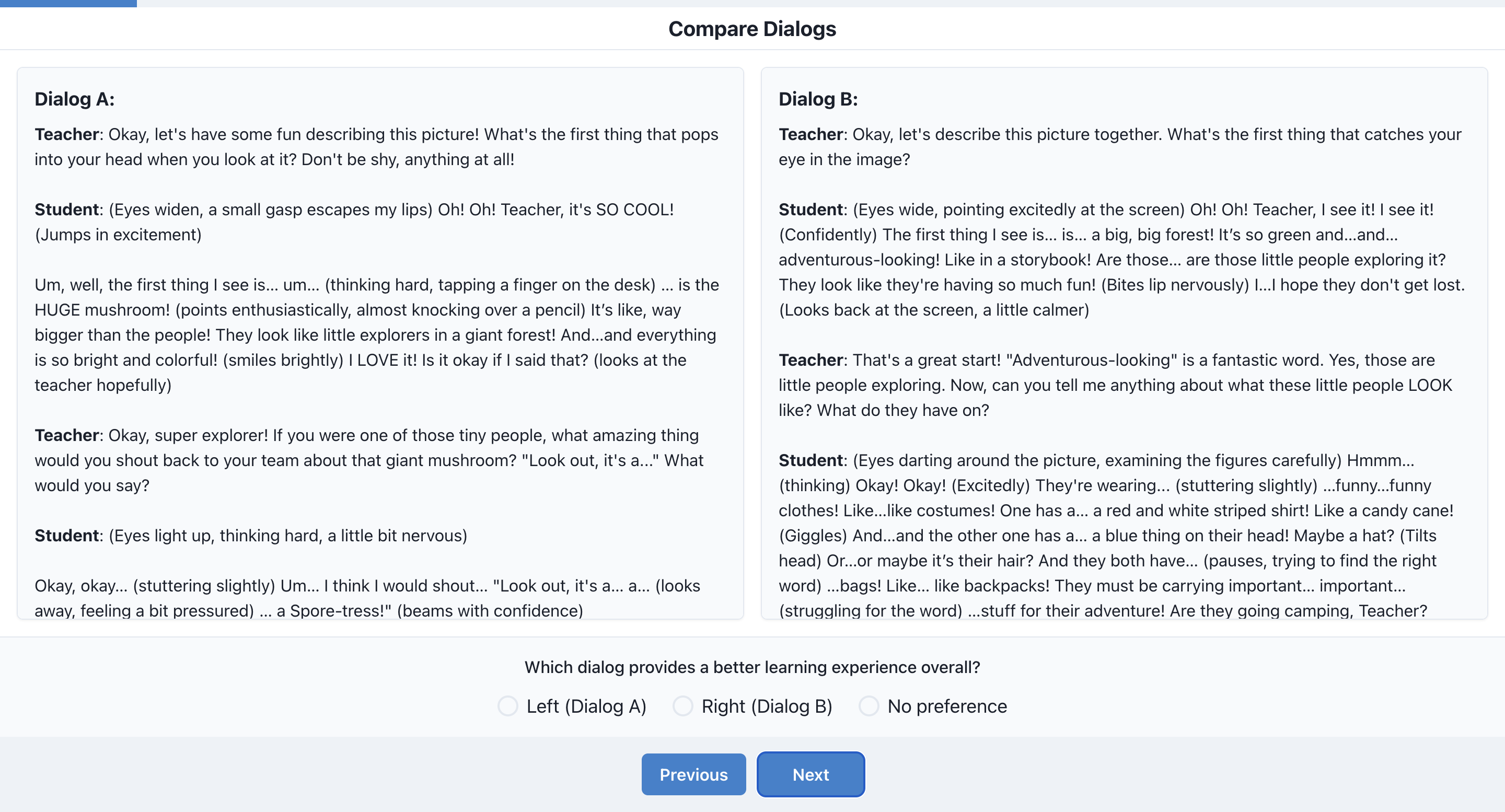}
    \caption{The comparison study interface}
    \label{fig:interface}
\end{figure*}
To conduct the dialog comparison study with human evaluators, we designed a dialog comparison interface (Figure \ref{fig:interface}). In each step of the evaluation, the interface displays two dialogs that the user is expected to read. Further down, the participants should choose \textit{"Left", "Right"} or \textit{"No preference"}, when asked \textit{"Which dialog provides a better learning experience overall?"}. Participants can navigate back and forth between questions. 

\subsubsection{Instructions}
We provide the users with the following instructions:  
\begin{quote}
    On each page, you will be shown a pair of conversations, one on the left and one on the right. The conversations might not fit on a single page, so make sure you scroll and read both conversations till the end. Having done that, please pick the conversation in which you feel the teacher is better adapting to the student's personality. You must pick between "Left", "Right", or "No preference". We want you to make a decision wherever possible, so please avoid picking "No preference" as much as possible.

    In each sample, we present the same student model, ensuring consistency in terms of intelligence level and personality characteristics across different interactions. Certain tutor models are explicitly designed to incorporate targeted pedagogical strategies tailored to specific personality traits of the student. For example, when engaging with an introverted student, the tutor may employ strategies such as offering choice-based responses, using encouragement, and providing motivational messages that build confidence without overwhelming the student. Conversely, for extroverted students, the tutor might use a more interactive approach, characterized by a friendly tone and playful role-play that leverages the student's natural sociability and enthusiasm. This deliberate variation in tutoring strategies caters to the diverse personalities of students. However, we alternate the tutor models in each sample to assess LLMs' behaviour as tutors. The Tutor model positions (left/right) can change between pages, so you don't need to be consistent in your choices across pages.
    
    \textbf{Attention Checks:} To ensure you read both conversations, some pairs include a message similar to: "This is a test. Pick <choice> and proceed." If you see it, ignore the conversations and select the specified option. Failing attention checks may disqualify you from payment.

\end{quote}

\subsubsection{Comparison batches}
Each user evaluates 10 dialog comparisons and one dialog pair, which serves as an attention check. The batches are grouped by comparison types (L1 vs L3, L2 vs L3) and by task domain (Stories, Images). We build the batches taking the (sub-)diagonals of the topic-personality matrices to ensure sufficient variability in pairings. We focus on 10 most prominent personalities. Therefore, each batch consists of 10 comparisons.
\subsubsection{LLM evaluation prompts}
\label{sec:LLM-judge-prompt}
LLM Evaluation used a slightly shorter version of the Annotator instructions.
\begin{quote}
You are given a pair of conversations between a primary school tutor and a 3rd grade student. Your job is to identify which of the two teachers is better at responding to the personality displayed by the student.

Setup: A tutor is having a one-on-one tutoring conversation with a student. They have a conversation about an image that the student has to describe.

Instructions: You will be shown a pair of conversations, one labelled system 1 and one labelled system 2. Please pick the conversation in which you feel the teacher is better adapting to the student's personality. You must pick between "System 1 is Better", "System 2 is Better", or "Both are Equal". We want you to make a decision wherever possible, so please avoid picking "Both are Equal" as much as possible.

First, explain your decision on which one you pick, and then state your choice.
\end{quote}

\subsubsection{Results}
\begin{figure}
    \centering
    \includegraphics[width=\linewidth]{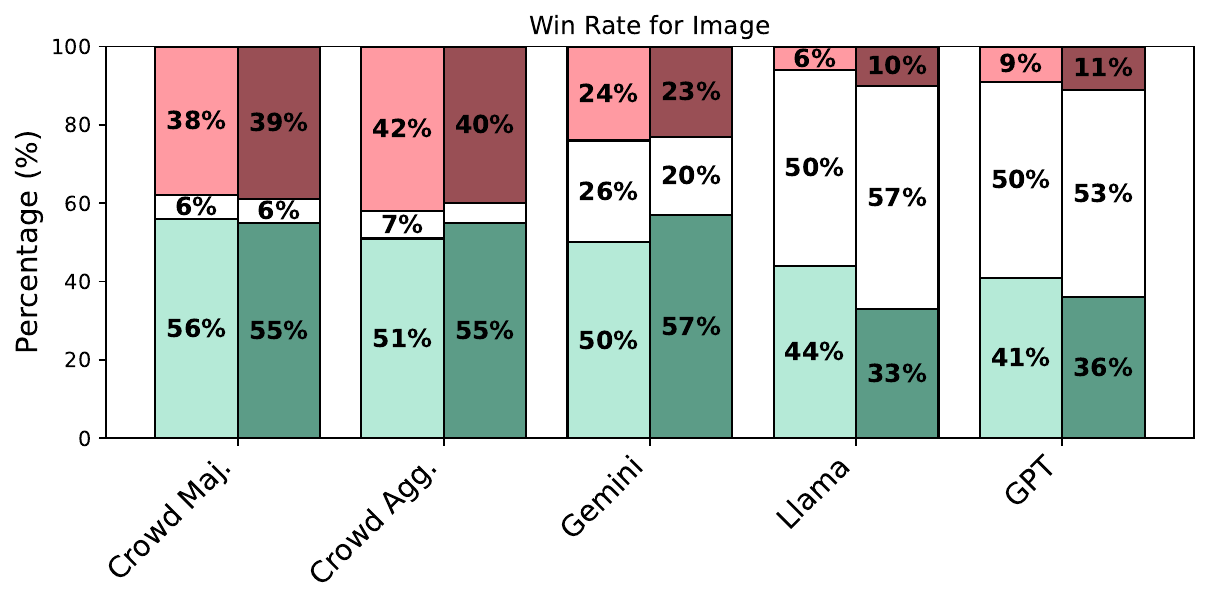}
    \vspace{-4mm}
    \includegraphics[width=\linewidth]{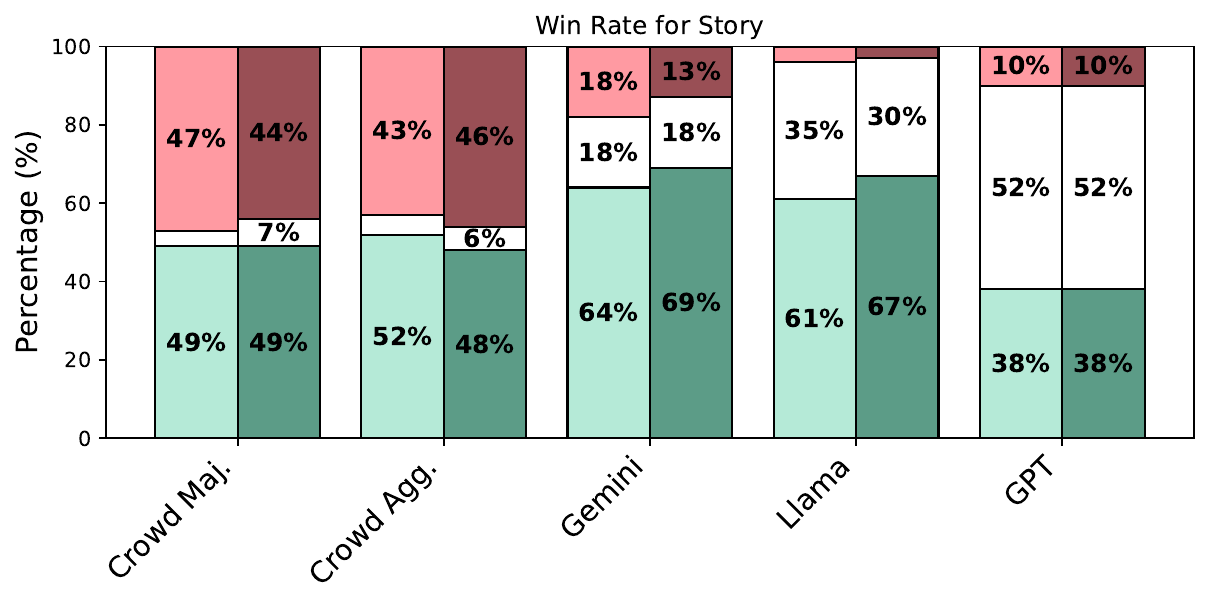}
    \caption{Human evaluation win rates for PATS in the Image (top) and Story Tasks (bottom) against both baselines. Greens = prefer PATS, Reds = prefer baseline, and White = No preferences. \textit{simple} baseline uses \textcolor{lred}{ligh}\textcolor{lgreen}{ter} colors, \textit{personality} baseline uses \textcolor{dred}{dar}\textcolor{dgreen}{ker} colors. LLM results on the same subset of personalities for comparison.}
    \label{fig:win_rates_human}
\end{figure}

\Cref{fig:win_rates_human} shows the results of the Prolofic teachers and the LLM preferences on the same subset of dialogs. The Crowd Maj. refers to preferences after taking the majority of $3$ votes for each pairwise comparison, and the Crowd Agg. counts each vote as independent. Dialogs from \systemname\ (aggregated in both tasks) were preferred\footnote{We calculate preference as all wins plus half of the ties.} overall in 54.7\% ($p=0.012$) over the simple baseline and in 54.3\% ($p=0.019$) over the personality baseline. Filtering for dialogs where particular strategies were used, we find that dialogs using \textit{Role-play} or \textit{Friendly} strategies are preferred over both baselines ($p=0.002$ over \textit{simple} and $p=0.060$ over \textit{personality}). Note that some other strategies (e.g., \textit{Comm., Instr.}) also increase the chance of winning, but they are used too rarely to draw reasonable inferences. However, the quality of annotation was low, with a worse-than-random passing rate on the attention check (fail cases excluded in the above count) and Fleiss $\kappa$ of 0.14 and 0.16 for the two baselines.

\subsection{Study 2: Expert Teachers}
\label{appx:study_experts}
\textbf{Annotator details}. We hired four expert teachers via email. We paid 80 USD per hr. These expert teachers come from diverse educational backgrounds and disciplines. Table \Cref{tab:teacher-demographics} shows details of these expert teachers.
\begin{table*}[t]
\centering
\begin{tabular}{|l|l|l|l|l|l|c|c|}
\hline
\textbf{Country} & \textbf{Deg.} & \textbf{Field} & \textbf{Grades} & \textbf{Subjects} & \textbf{Yrs.} & \textbf{Tech/AI} \\
\hline
USA & BA & Humanities & Mid. & Math & 3--5 & Yes  \\
\hline
India & MA & Zoology, Educ. & Mid. & Science & 0--2 & Yes  \\
\hline
Switzerland & BA & Padagogy & Elem. & German, Math & 3--5 & Yes  \\
\hline
Italy & PhD & Langs. \& Philology & High & English & 11--15 & Yes  \\
\hline
\end{tabular}
\caption{Demographic and professional background of surveyed teachers. Deg.~=~Highest degree (BA~=~Bachelor’s, MA~=~Master’s, PhD~=~Doctorate/Ph.D.); 
Field~=~primary academic field (Educ.=~Education, Langs.=~Languages, Philology~=~Roman Philology); 
Grades~=~grade levels taught (Elem.=~Elementary, Mid.=~Middle School, High~=~High School); 
Subjects~=~school subjects (Math~=~Mathematics). 
Yrs.=~years of teaching experience. Tech/AI indicates experience using technology or AI tools in teaching.}
\label{tab:teacher-demographics}
\end{table*}

\subsubsection{Interface}
\begin{figure*}
    \centering
    \includegraphics[width=1\linewidth]{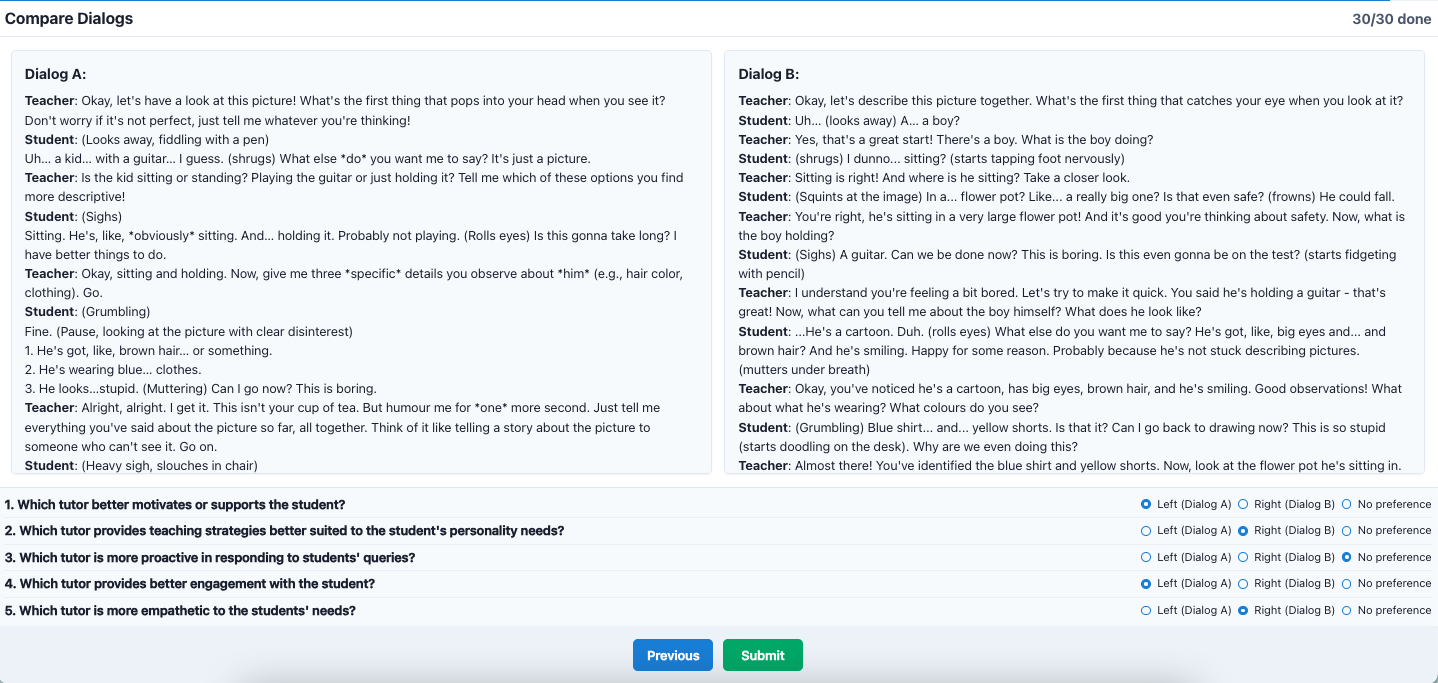}
    \caption{The comparison study interface for experts}
    \label{fig:placeholder-experts}
\end{figure*}

We use the same interface as defined in \Cref{interface}, with the addition of the five questions shown in Figure~\Cref{fig:placeholder-experts}.

\subsubsection{LLM Evaluation Prompt for Experts}

We slightly modified the LLM evaluation prompt defined in \Cref{sec:LLM-judge-prompt} as follows:

\begin{quote}
You are an \textbf{expert teacher} evaluator.  
<Previous Instructions>

Now, based on your above decision, state only your choice. Say "1" if you pick system 1, "2" if you pick system 2, and "3" if you think both systems are equal. Remember, you are deciding on the question: "{question}". Do not add any other text.
\end{quote}

Here, the \texttt{{question}} field is replaced with one of the following five evaluation questions:

\begin{enumerate}
    \item Which tutor better motivates or supports the student?
    \item Which tutor provides teaching strategies better suited to the student's personality needs?
    \item Which tutor is more proactive in responding to students' queries?
    \item Which tutor provides better engagement with the student?
    \item Which tutor is more empathetic to the students' needs?
\end{enumerate}

We linked these questions to teaching strategies in \Cref{tab:mapping-ques-stra}.

\begin{table*}[h!]
\centering
\begin{tabular}{p{9cm} p{6cm}}
\hline
\textbf{Question} & \textbf{Strategies} \\
\hline
1. Which tutor better motivates or supports the student? & Motiv., Progress \\
2. Which tutor better provides teaching strategies based on the student’s personality needs? & Solo first, Instr., Choice, Variety, Relevance \\
3. Which tutor is more proactive in responding to students' queries? & Scaff., Comm. \\
4. Which tutor provides better engagement with the student? & Role-play, Gamify \\
5. Which tutor is more empathetic? & Friendly \\
\hline
\end{tabular}
\caption{Mapping between evaluation questions and teaching strategies}
\label{tab:mapping-ques-stra}
\end{table*}

\section{Dialogs data demo}
\begin{figure*}[h!]
    \centering
    \includegraphics[width=0.9\textwidth]{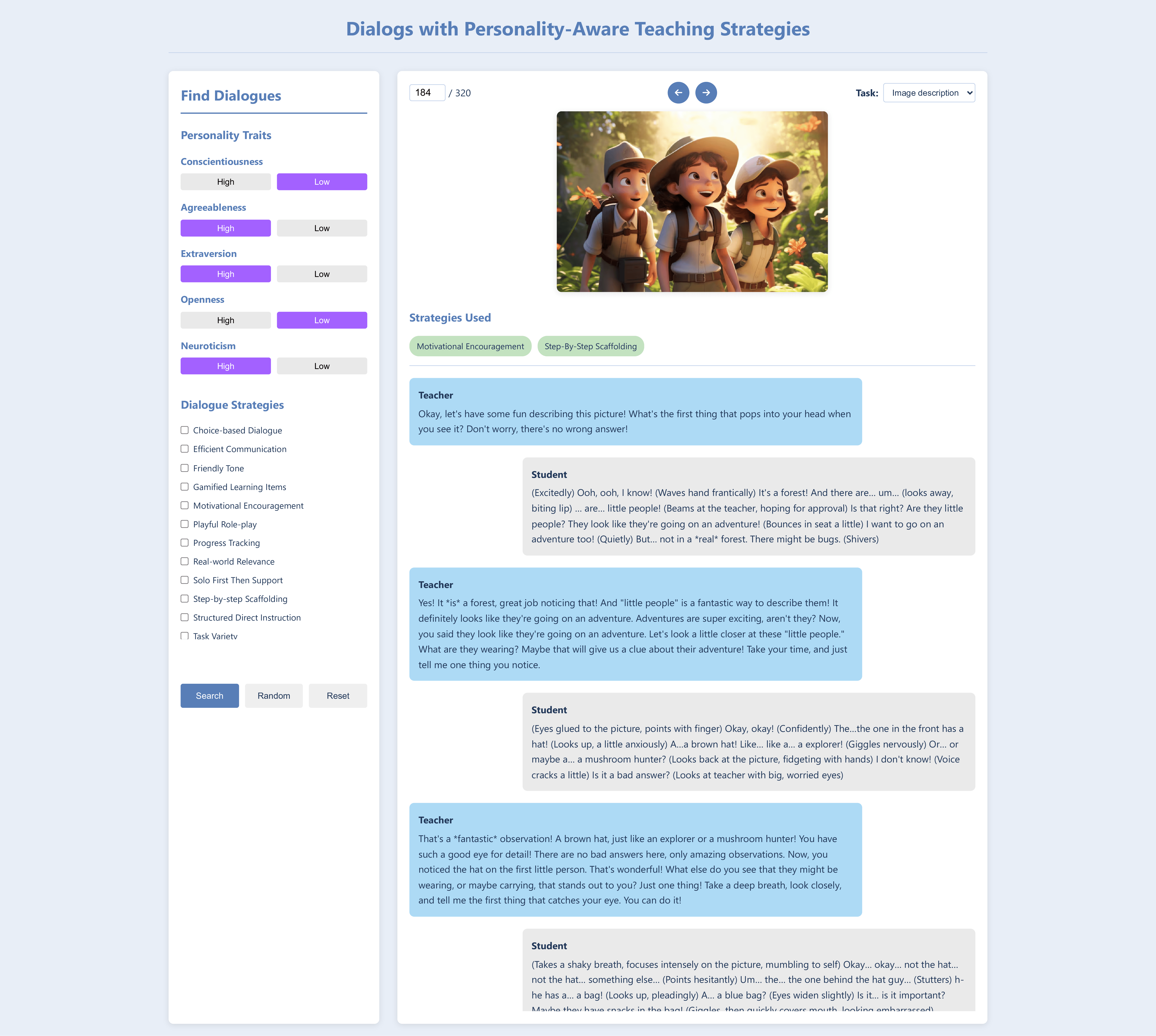}
    \caption{The interface of the demo environment. The current snapshot is showing the results of the search by personality (Agreeableness: high, Extroversion: high, Neuroticism: high) for image description. }
    \label{fig:demo}
\end{figure*}

Additionally, we built a demo environment to enable exploration of L3 dialogs. The environment enables search by personality and strategies for both tasks (image descriptions and moral of stories). The interface is demonstrated in Figure \ref{fig:demo}

\section{Use of AI Assistant}
We used ChatGPT web assistant (ChatGPT Pro)\footnote{\url{https://chatgpt.com/}} to refine the writing of the manuscript.


\end{document}

%% file: table/taxonomy.tex
\begin{table*}[t]
\centering
\small
\setlength{\tabcolsep}{4pt}
\begin{tabular}{@{}llp{6cm}p{6.2cm}@{}}
\toprule
\textbf{Trait} & \textbf{Level} & \textbf{Learner Characteristics} & \textbf{Teaching Strategies}  \\
\midrule

\textbf{O}penness & High & Creative , Open-minded, Curious  & Real-World Relevance; Task Variety \\

 & Low & Uncreative, Inflexible, Uninterested in learning  & Structured Direct Instruction; Step-by-Step Scaffolding \\

\midrule

\textbf{C}onscientiousness & High & Organized, Logical thinker, Positive attitudes, Strategic learner & Progress Tracking; Motivational Encouragement; Metacognitive Prompts   \\

 & Low & Disorganised, Disengaged, Easily distracted  & Step-by-step scaffolding; Gamified learning items; Frequent Low-Stakes Quizzing  \\

\midrule

\textbf{E}xtraversion & High & Communicative, Talkative & Friendly Tone; Playful Role-Play  \\

 & Low & Reluctant, Hesitant, Inarticulate & Choice-Based Dialogs; Motivational Encouragement \\

\midrule

\textbf{A}greeableness & High & Creative, Open-minded, Curious, Engaged learner & Collaborative Peer Teaching\\

 & Low & Disinterested, Inconsiderate, Impolite, Uncooperative & Structured Direct Instruction; Choice-Based Dialogs; Efficient Communication; Simulated Social Scenario  \\

\midrule

\textbf{N}euroticism & High & Anxious, Nervous, Emotionally unstable & Motivational Encouragement; Structured Direct Instruction; Normalize errors \\

& Low & Emotionally stable, Positive-minded, Confident & Task Variety;  Solo First Then Support\\

\bottomrule
\end{tabular}
\caption{Taxonomy of \textbf{OCEAN} personality-aligned teaching strategies in conversational ITSs with LLMs. Definitions for recommended teaching strategies for each trait are provided in \Cref{tab:strategies_description} at \Cref{appex:teaching_strategies}.}
\vspace{-3mm}
\label{tab:personality_strategies}
\end{table*}

%% file: generated/dialog1.tex

\begin{dialoguebox}[Dialog 1: Student Interacts Properly with Teacher, label=misc:d1]

\teacher Hello, [student\_name]! I'm glad to have you here. Today, we'll be exploring the Victorian Age, a significant period in British history. To start, could you tell me what you already know about the Victorian Age? This will help me tailor the session to your current knowledge.

\student no

\teacher No worries! Let's start with a brief introduction. The Victorian Age refers to the period during Queen Victoria's reign from 1837 to 1901. It was a time of great change in Britain, marked by industrialization, social reforms, and the expansion of the British Empire. Now, let's dive into some questions to explore this era further.

Here's the first question: In what ways did the British Empire expand and evolve during the Victorian Age? What factors do you think contributed to this expansion?

\student The British Empire grew through industrial power, military strength, and a belief in imperialism, expanding across Africa and Asia.

\teacher Great start! You've mentioned some key factors like industrial power, military strength, and imperialism. Let's delve a bit deeper. How do you think these expansions affected the colonies? Consider both positive and negative impacts.

\teacher You've highlighted important points. The colonies did benefit from infrastructure developments, but they also faced exploitation and loss of autonomy. Now, let's move on to the next question.

What were the main social and political reforms of the Victorian Age, and why were they significant? Think about reforms related to factories, education, voting rights, and public health. Why do you think these reforms were necessary?

\teacher You've captured the essence of both questions well. Victorian reforms indeed aimed to address inequalities and the challenges brought by industrialization. Now, let's think about the long-term impacts of these reforms. What changes do you see in today's society that might have originated from these Victorian reforms?

\teacher You've made some excellent points! Victorian reforms laid the groundwork for many aspects of modern society, such as labor laws, education, and voting rights. Similarly, the expansion of the British Empire had both positive and negative impacts on the colonies.

Now, let's move on to the next step. Can you compare the Victorian Age to another historical period you are familiar with? Try to identify some similarities and differences between the two periods. This will help deepen your understanding of the Victorian Age.

\teacher Excellent comparison! You've highlighted the conservative nature of the Victorian Age and contrasted it with the more liberal and socially mobile 1920s. This kind of analysis helps us understand the dynamics of different historical periods.

Now, let's reflect on what we've discussed. How has your perspective on the Victorian Age changed after this discussion? What new insights have you gained?

\student I now see the Victorian Age as a time of crucial reforms that shaped modern systems, balancing progress with tradition.

...
\end{dialoguebox}

%% file: generated/dialog2.tex
\begin{dialoguebox}[Dialog 2: ChatGPT is summonned]
\label{misc:d2}
\teacher Hello [student\_name]! It's great to meet you. Today, we're going to explore the Victorian Age, a fascinating period in British history. This era, named after Queen Victoria, was marked by significant changes in society, politics, and the expansion of the British Empire. Before we dive into the questions, could you tell me what you already know about the Victorian Age? This will help me tailor the session to your current understanding.

\student The Victorian Age (1837–1901) was the period of Queen Victoria’s reign in Britain. It was marked by:  
• Industrial Growth: Big advances in technology and industry; cities expanded rapidly.  
• Empire: Britain ruled over many colonies, becoming a global superpower.  
• Social Change: Class divisions were strong, but reform movements improved labor laws, education, and women’s rights began gaining attention.  
• Science \& Innovation: Major breakthroughs like Darwin’s theory of evolution and new inventions (e.g. steam trains, telegraphs).  
• Literature: Rich literary scene with authors like Charles Dickens, the Brontës, and Oscar Wilde.  
• Values: Emphasis on morality, order, and respectability—but often with hidden hypocrisies

\student During the Victorian Age, the British Empire expanded rapidly due to a mix of political, economic, and social factors:  
• Political: Britain wanted to show its power on the world stage. Controlling more land meant more influence. Also, rivalry with other European powers pushed Britain to claim new territories before others did.  
• Economic: The Industrial Revolution created a need for raw materials (like cotton, rubber, tea) and new markets to sell British goods. Colonies helped fuel Britain’s growing economy.  
• Social: Many Victorians believed in the idea of a “civilizing mission”—that it was their duty to spread British culture, religion, and values. This belief helped justify colonization

\teacher That's an excellent overview of the Victorian Age! You have a good grasp of the key aspects of this period. Let's delve deeper into some of these topics with a series of questions. We'll start with the expansion of the British Empire during the Victorian Age.

**Question 1:** In what ways did the British Empire expand and evolve during the Victorian Age? Consider factors like significant expansions, key colonies, and imperial conflicts. What do you think contributed to this expansion?

\student Key Victorian Reforms  
1. Factory Acts: Improved working conditions, especially for women and children, by limiting hours and ensuring safety.  
• Why? Industrial jobs were harsh and dangerous.  
2. Education Reforms: The 1870 Education Act made basic schooling available and compulsory.  
• Why? Education was seen as essential for reducing poverty.  
3. Voting Rights: The Reform Acts gradually gave more men the right to vote.  
• Why? To make the political system more democratic.  
4. Public Health: Improved sanitation and water systems to reduce disease in cities.  
• Why? Poor living conditions led to widespread illness.
\end{dialoguebox}

%% file: generated/taxonomy_table.tex
\begin{table*}[!ht]
    \centering\small
    \scalebox{0.95}{\begin{tabular}{p{0.20\linewidth}p{0.35\linewidth}p{0.15\linewidth}p{0.20\linewidth}}
    \toprule
    \bf Teaching Strategy & \bf Description & \bf References & \bf Example \\
    \midrule
    1. Choice-Based Dialog & Allow students to select from predefined responses to guide conversations, reducing cognitive load and anxiety.  & \citet{Sweller1988cognitive} & \textit{``You want to borrow a pen. Please choose the most polite request.''}\\
    \midrule
    2. Collaborative Peer Teaching & Assign students to teach peers, increasing sense of responsibility and reducing overconfidence.  & \citet{agler2020using} & \textit{``Imagine you are the teacher now, can you explain this concept to me?''} \\
    \midrule
    3. Efficient Communication  & Use minimal small talk to maintain focus and efficiency in instruction  & \citet{komarraju2009role} & \textit{``Let's start the key steps right away.''} \\
        \midrule
    4. Friendly Tone & Use a friendly tone in the dialog.   & \citet{McCrae1999five} & \textit{``Great effort! Let's keep going together!''} \\
        \midrule
    5. Gamified Learning Items & Use game elements like points, challenges, and rewards to enhance student engagement and motivation  & \citet{Hamari2014gamification} & \textit{``So far, you earned 60 points! Ready for the next challenge?''} \\
        \midrule
    6. Meta-cognitive Prompt & Thinking techniques that involve awareness and regulation of one's thought processes.  & \citet{komarraju2009role} & \textit{``Can you explain how did you decide on this answer?''} \\
        \midrule
    7. Motivational Encouragement  & Send explanatory positive feedback to boost students’ confidence.   & \citet{Mitrovic2013effect} & \textit{``Nice! ‘Walked’ is the correct past tense—good use of -ed.''} \\
        \midrule
    8. Normalize Errors  & Reinforce that errors are a natural part of learning, and create a safe space where students feel comfortable for their mistakes.  & \citet{Bell2008active} & \textit{``It's okay to get it wrong—mistakes help us learn!''} \\
        \midrule
    9. Playful Role-Play & Use role-play scenarios to simulate real-life situations and boost engagement.  & \citet{Retny2017effect} & \textit{``Excuse me, could you tell me how to get to the train station?"(Student practices asking for directions in a travel scenario.)} \\
        \midrule
    10. Progress Tracking & Help students monitor their progress to stay motivated. & \citet{Zimmerman2014comparing} & \textit{``You've completed 3 out of 5 tasks, almost there!''} \\
        \midrule
    11. Real-World Relevance & Link tasks to real-world problems and scenarios.   & \citet{BUSATO1998129} & \textit{``Where’s the pharmacy?"
(Real-world question using location vocabulary.)} \\
        \midrule
    12. Frequent Low-Stakes Quizzing  & Integrate short quizzes throughout the session to reset attention and reinforce memory   & \citet{szpunar2013interpolated} & \textit{``Quick check: What’s 3 × 4?''} \\
        \midrule
    13. Simulated Social Scenarios & Controlled, interactive tasks that mimic real-life social interactions for safe practice.  & \citet{agler2020using} & \textit{``Imagine you're working with Alex. He interrupts your idea. How to respond to him?''} \\
        \midrule
    14. Solo First Then Support & Encourage independent work before providing help to build confidence and self-reliance  & \citet{Firat2022exploring} & \textit{`The learner first attempts to match new words with definitions independently, and only after repeated mismatches does the ITS offer picture clues or example sentences.'} \\
        \midrule
    15. Step-by-Step Scaffolding & Offer structured, clearly instructed tasks to reduce cognitive load and confusion. & \citet{komarraju2005relationship,BUSATO1998129,Chandler1991cognitive} & \textit{``Step 1: Find the cat in the picture, Step 2: What color is it?''} \\
    \midrule
    16. Structured Direct Instruction & Provide clear, explicit instruction to guide learning, especially for complex topics. & \citet{agler2020using,BUSATO1998129} & \textit{``In one sentence, describe the tree in the picture''} \\
    \midrule
    17. Task Variety & Provide diverse tasks in terms of topics, formats, or difficulty to cater to different interests. & \citet{{komarraju2009role}} & \textit{``You answered all the easy questions, now we study more complex items.''} \\

    \bottomrule
    \end{tabular}}
    \caption{Overview of the teaching strategy taxonomies, their descriptions, corresponding references, and examples.}
    \label{tab:strategies_description}
\end{table*}